\documentclass[10pt,twocolumn,letterpaper]{article}

\usepackage{iccv}

\usepackage[utf8]{inputenc} 
\usepackage[T1]{fontenc}    
\usepackage{lscape}
\usepackage{graphicx}
\usepackage{amsmath}
\usepackage{amssymb}
\usepackage{booktabs}
\usepackage{enumitem}
\usepackage{algorithm}
\usepackage{algorithmic}
\usepackage{float} 
\usepackage{acronym}
\usepackage{balance}
\usepackage{xspace}
\usepackage{setspace}
\usepackage{tabularx,colortbl,multirow,array,makecell}
\usepackage{overpic,wrapfig}
\usepackage{nicefrac}
\usepackage[misc]{ifsym}
\usepackage{siunitx}
\usepackage{pifont}
\definecolor{cvprblue}{rgb}{0.21,0.49,0.74}
\usepackage[pagebackref,breaklinks,colorlinks,allcolors=cvprblue]{hyperref}
\usepackage[capitalize]{cleveref}
\usepackage{flushend}

\definecolor{gblue}{HTML}{4285F4}
\definecolor{gred}{HTML}{DB4437}
\definecolor{ggreen}{HTML}{0F9D58}
\definecolor{vblue}{HTML}{2993ba}

\newcommand{\model}{\textbf{\mbox{TACO}}\xspace}
\newcommand{\supp}{\textit{supplementary materials}\xspace}

\acrodef{vac}[VAC]{Video Amodal Completion}
\acrodef{vas}[VAS]{Video Amodal Segmentation}
\acrodef{ovo}[OvO]{Object-video-Overlay}

\makeatletter
\renewcommand{\paragraph}{%
  \@startsection{paragraph}{4}{\z@}%
  {1ex plus 0.5ex minus 0.2ex} 
  {-1em}                      
  {\normalfont\normalsize\bfseries} 
}
\makeatother

\title{TACO: \underline{T}aming Diffusion for in-the-wild Video \underline{A}modal \underline{Co}mpletion}

\author{
  Ruijie Lu\textsuperscript{1,2\,$\star{}$},
  Yixin Chen\textsuperscript{2$\dagger$},
  Yu Liu\textsuperscript{2,3},
  Jiaxiang Tang\textsuperscript{1}, \\
  Junfeng Ni\textsuperscript{2,3}, 
  Diwen Wan\textsuperscript{1},
  Gang Zeng\textsuperscript{1$\dagger$},
  Siyuan Huang\textsuperscript{2$\dagger$}
  \\
  \small \textsuperscript{$\star{}$} Work done as an intern at BIGAI \,\textsuperscript{$\dagger$} Corresponding Authors
  \\
  \small \textsuperscript{1} State Key Laboratory of General Artificial Intelligence, Peking University \\
  \small \textsuperscript{2} State Key Laboratory of General Artificial Intelligence, BIGAI 
  \textsuperscript{3} Tsinghua University
}

\begin{document}

\twocolumn[{
\renewcommand\twocolumn[1][]{#1}
\maketitle
\vspace{-0.4in}
\begin{center}
    \centering
    \captionsetup{type=figure}
    \includegraphics[width=0.95\linewidth]{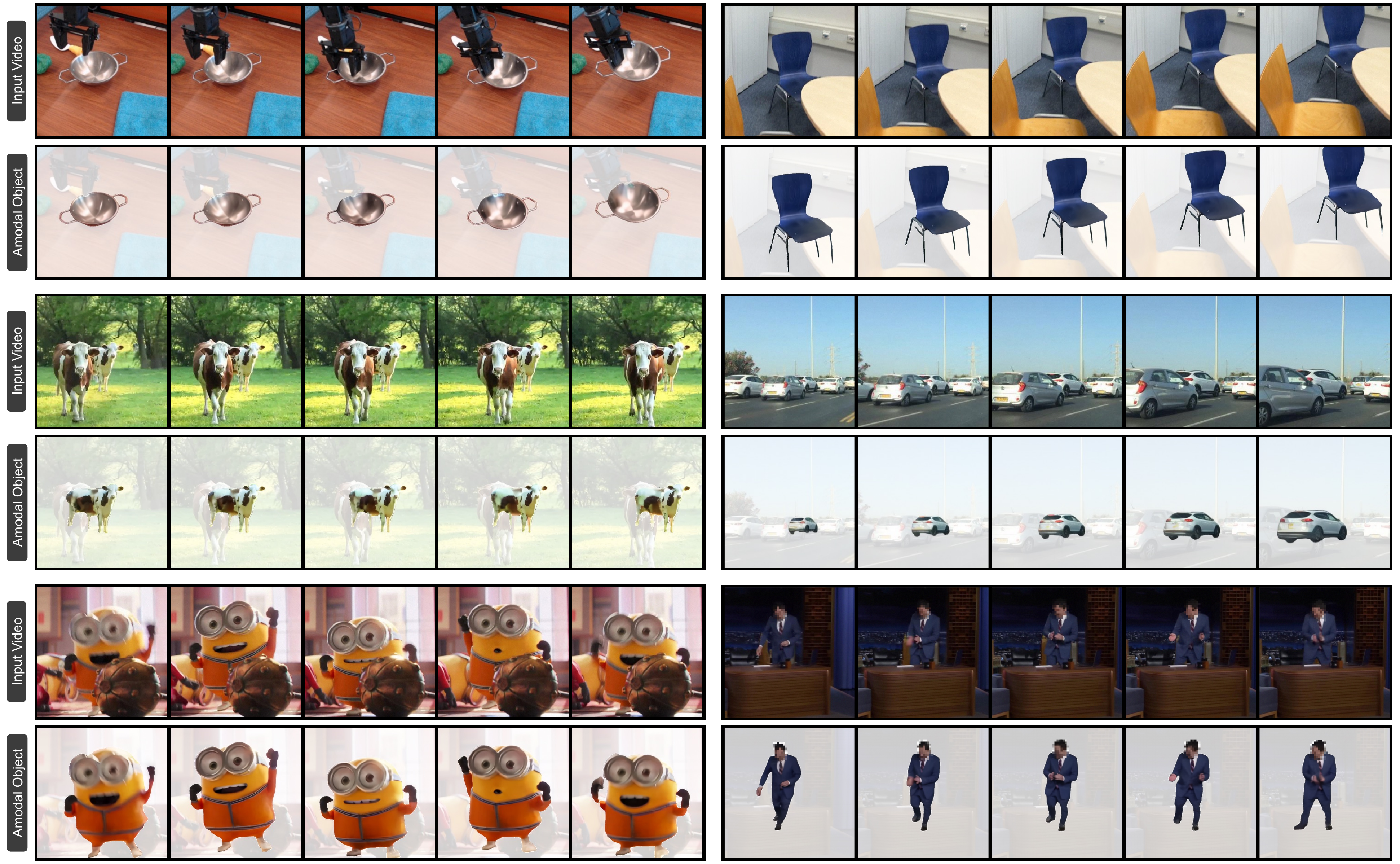}
    \captionof{figure}{
        We propose \model for video amodal completion. Our model is capable of synthesizing consistent amodal contents in diverse in-the-wild occluded scenarios, including robotic manipulation, scene understanding, autonomous driving, and Internet videos. 
    }
    \label{fig:teaser}
\end{center}
}]

\begin{abstract}
Humans can infer complete shapes and appearances of objects from limited visual cues, relying on extensive prior knowledge of the physical world. However, completing partially observable objects while ensuring consistency across video frames remains challenging for existing models, especially for unstructured, in-the-wild videos.
This paper tackles \ac{vac}, aiming to generate the complete object consistently throughout the video given a visual prompt specifying the object of interest. Leveraging the rich, consistent manifolds learned by pre-trained video diffusion models, we propose a conditional diffusion model, \model, that repurposes these manifolds for VAC. To enable its effective and robust generalization to challenging in-the-wild scenarios, we curate a large-scale synthetic dataset with multiple difficulty levels by systematically imposing occlusions onto un-occluded videos. Building on this, we devise a progressive fine-tuning paradigm that starts with simpler recovery tasks and gradually advances to more complex ones. We demonstrate \model's versatility on a wide range of in-the-wild videos from Internet, as well as on diverse, unseen datasets commonly used in autonomous driving, robotic manipulation, and scene understanding. Moreover, we show that \model can be effectively applied to various downstream tasks like object reconstruction and pose estimation, highlighting its potential to facilitate physical world understanding and reasoning. Our project page is available at \href{https://jason-aplp.github.io/TACO/}{https://jason-aplp.github.io/TACO/}.
\end{abstract}

\section{Introduction}
\label{sec:intro}

Humans excel at inferring complete shapes and appearances of partially observable objects, as explained by the continuity and closure principle from Gestalt psychology~\cite{wertheimer1912experimentelle,wagemans2012century,koffka2013principles,wertheimer2017untersuchungen}. This capability is challenging for current models as it requires consistently completing a target object across frames and generalizing to diverse in-the-wild videos, due to the innate ambiguities in unobserved areas.
Such consistent amodal representations are crucial for physical world understanding, supporting reasoning~\cite{locatello2020object} and facilitating downstream tasks like reconstruction~\cite{wang2024dust3r, chen2023ssr, yu2022monosdf, liu2024slotlifter, wang2024roomtex, ni2024phyrecon, wan2024superpoint, ni2025dprecon, shen2025trace3d}, robotics~\cite{li2024ag2manip, brohan2023rt, team2024octo, zhu2023viola, liu2025building, lu2024manigaussian}, autonomous driving~\cite{ma2019accurate, chiu2021probabilistic}, and beyond~\cite{huang2025leo, huang2025unveiling, huang2023embodied}. Therefore, it is vital to develop a model's ``consistent see through'' capability, especially when applied to diverse in-the-wild videos.

In this paper, we tackle the task of \Acf{vac}, recovering spatio-temporal consistent and complete objects given occluded observations in a video. 
Previous work mainly focuses on related but distinct aspects of visual content completion. Image amodal completion~\cite{ozguroglu2024pix2gestalt, xu2024amodal, zhan2024amodal} and inpainting~\cite{lugmayr2022repaint, tang2024intex, ju2024brushnet} can be extended to videos by processing each frame independently, but they ignore neighboring frames and fail to predict amodal objects consistently. 
Video inpainting~\cite{gao2020flow, ke2021occlusion, kim2019deep, lee2019copy, li2022towards, zeng2020learning} can also be adapted to this problem. However, these methods require specifying masks, implying the complete shape of occluded objects. Additionally, they do not possess strong object awareness, resulting in obvious artifacts when completing occluded objects in videos~\cite{li2024ag2manip}.


Building on advances in video generation~\cite{blattmann2023stable, hong2022cogvideo, xing2025dynamicrafter, guoanimatediff}, we repurpose pre-trained models, which have learned rich spatio-temporally consistent manifolds from Internet-scale datasets~\cite{bain2021frozen}, for amodal completion. We introduce \model, a conditional video diffusion model, that given a video and a visual prompt specifying an object, generates its consistent amodal representation across frames.

Our key insight lies in how to effectively fine-tune the model for better generalization and robustness across diverse, challenging scenarios.
Simulator data~\cite{hu2019sail} inherently suffers from a sim-to-real gap, while real-world videos lack ground-truth amodal references for occluded objects. Inspired by pix2gestalt~\cite{ozguroglu2024pix2gestalt}, we utilize real-world videos and curate a diverse, scalable synthetic dataset by identifying unoccluded objects in videos and then overlaying occluders consistently. As shown in SAM2~\cite{ravi2024sam}, training on increasingly more challenging data is beneficial for generalization. Thus, we adopt a \textit{model-in-the-loop} approach that identifies failures and difficult data samples to guide subsequent data augmentation. This results in roughly 200k video pairs at varying difficulty levels. We then employ a progressive training strategy, beginning with simpler recovery tasks and gradually advancing to more complex ones, which is critical for robust generalization to challenging in-the-wild videos.

Extensive experiments demonstrate that our method can synthesize consistent amodal objects in shape, appearance, and motion. It generalizes well to unseen datasets like BridgeData~\cite{ebert2021bridge, walke2023bridgedata}, ScanNet~\cite{yeshwanth2023scannet++, dai2017scannet} and in-the-wild Internet videos, as shown in \cref{fig:teaser}. Our model also achieves state-of-the-art performance on \ac{vas} in a zero-shot manner. Beyond the model's ``see-through'' capability, we show our model can directly facilitate downstream tasks as a drop-in module in object reconstruction and 6-DoF pose estimation.

To sum up, our main contributions are:
\begin{enumerate}[nolistsep,noitemsep,leftmargin=*]
    \item We propose \model, a conditional video diffusion model for \acf{vac}, enabling spatio-temporally consistent recovery of amodal objects from monocular video in occluded scenarios.
    \item We present a scalable data curation pipeline for constructing occluded and unoccluded video pairs as well as a progressive training paradigm for effective and robust generalization to unseen in-the-wild scenarios.
    \item Experiments show that our approach outperforms all previous methods in \acl{vac} and performs on par with SOTA in \acl{vas} in a zero-shot manner. Furthermore, we directly leverage our model to improve downstream tasks like object reconstruction and 6-DoF pose estimation, highlighting its potential in real-world applications.
\end{enumerate}


\section{Related Work}
\label{sec:related_work}
\vspace{-0.2em}
\paragraph{Amodal Segmentation and Completion}
Amodal completion is challenging due to the ambiguity in occluded regions. Most previous works focus on simpler tasks, such as amodal segmentation~\cite{ke2021deep, ling2020variational, qi2019amodal, reddy2022walt, zhu2017semantic} or detection~\cite{kar2015amodal}, predicting amodal segmentation masks or bounding boxes. Follow-up studies~\cite{hsieh2023tracking, yao2022self, fan2023rethinking, hu2019sail, van2023tracking, dave2020tao, athar2023burst} extend amodal segmentation to the video domain to ensure consistency. However, these approaches do not generate content for occluded areas. More recent works~\cite{ozguroglu2024pix2gestalt, xu2024amodal, zhan2024amodal} attempt to unleash the capabilities of pre-trained diffusion models~\cite{rombach2022high} to generate content for these areas. 
Yet, these models are limited to the image level, making it difficult to maintain spatial-temporal consistency and utilize information from neighboring frames, which often reveal occluded parts from alternative viewpoints and provide valuable context for generating coherent content.
Instead, we propose to address this task directly within the video domain, enabling a broader receptive field across the entire video and synthesizing more consistent results. Concurrent work~\cite{chen2024using} also explores video amodal completion in two stages by first predicting amodal masks; however, its reliance on simulation data, \ie, SAIL-VOS~\cite{hu2019sail}, and the prerequisite of pseudo-depth limit its applicability in more generalized use cases.

\begin{figure*}[ht!]
    \centering
    \includegraphics[width=0.95\linewidth]{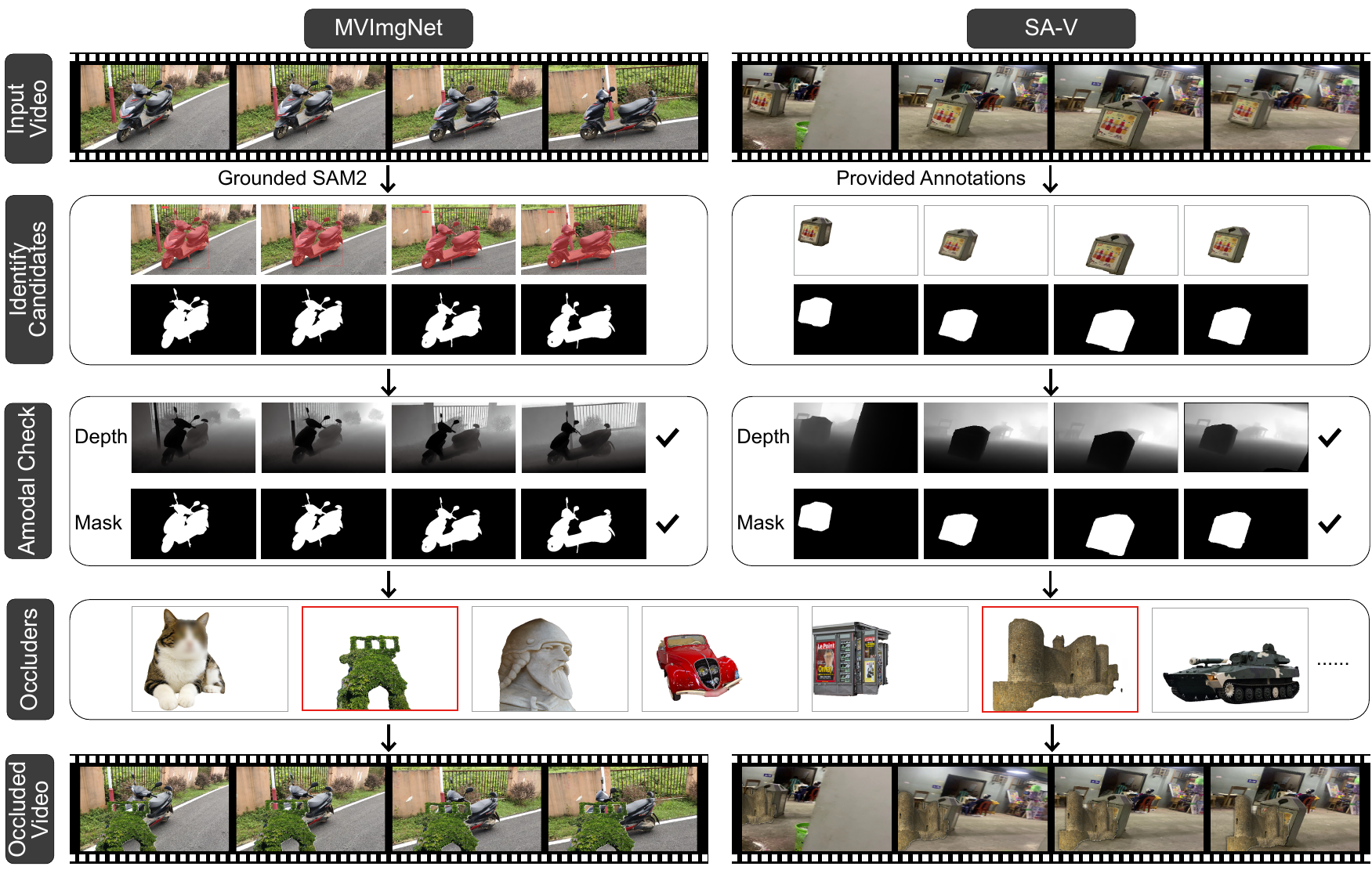}
    \caption{\textbf{Data curation pipeline.} We generate our synthetic custom dataset following three steps. First, given an input video, we identify possibly unoccluded candidates using an off-the-shelf segmentation model~\cite{ren2024grounded} or provided annotations. Next, we check whether the candidate is amodal with heuristic rules and manual filtering. Finally, we select an occluder and overlay it upon the candidate consistently. }
    \label{fig:data_pipeline}
    \vspace{-1.8em}
\end{figure*}

\paragraph{Video Inpainting}
Video inpainting, which fills masked areas with plausible and consistent content, can also be repurposed for \acf{vac}. Early explorations mainly focus on how to transfer information from neighboring frames to the
target frame~\cite{quan2024deep} using 3D convolutions~\cite{hu2020proposal, chang2019learnable}, flow~\cite{gao2020flow, xu2019deep, li2022towards}, or cross-attention~\cite{lee2019copy, li2020short, liu2021fuseformer}. More recently, research work~\cite{zhang2024any, wu2024towards} proposes to incorporate text prompts when inpainting a video, leveraging powerful text-based image inpainting techniques~\cite{lugmayr2022repaint, corneanu2024latentpaint}. However, all these inpainting-based methods require specifying masked regions, which is non-trivial since we do not know the occluded object's shape. Furthermore, video inpainting models are proposed for more general visual completion; they lack amodal object awareness, making it difficult to accurately recover occluded portions of objects, even when provided with a precise inpainting mask~\cite{li2024ag2manip}.

\paragraph{Video Diffusion Models}
We have witnessed huge advances in video generation~\cite{blattmann2023stable, blattmann2023align, ge2023preserve, ho2022imagen, singer2022make, xing2025dynamicrafter, hong2022cogvideo, yang2024cogvideox, guoanimatediff, agarwal2025cosmos} recently, with most approaches being diffusion-based, building on the success of text-to-image diffusion models~\cite{rombach2022high, ramesh2021zero, saharia2022photorealistic}. 
Leveraging the rich and consistent manifolds learned by large-scale video diffusion models, subsequent works have explored repurposing these pre-trained models for novel tasks such as depth estimation~\cite{shao2025learning, hu2024depthcrafter, yang2024depth, ke2023repurposing, gui2025depthfm}, novel view synthesis (NVS)~\cite{van2024generative, yu2024viewcrafter, liu2024reconx, liu2023zero, lu2024movis}, assets generation~\cite{voleti2025sv3d, xie2024sv4d, zuo2024videomv}, and so on~\cite{li2024puppet, liang2024dreamitate}. In this work, we tackle the challenging task of video amodal completion. We use publicly available Stable Video Diffusion (SVD)~\cite{blattmann2023stable}, introducing a scalable dataset and employing a progressive training paradigm for effective fine-tuning.

\section{Data Curation}
\label{sec:method_data}
For the training and evaluation of the \ac{vac} task, we curate a large-scale dataset, \acf{ovo}. Ideally, a \ac{vac} dataset should comprise diverse real-world videos containing occluded objects alongside their unoccluded versions. While such data pairs can be scalably generated in simulated environments~\cite{blender, xiang2020sapien}, relying solely on simulator data inevitably introduces a sim-to-real gap~\cite{hofer2021sim2real}, limiting generalization capability. Contrarily, collecting such data pairs at scale in the real world is costly and impractical, as it requires removing occluders while replicating the motion of both the camera and other dynamics within a scene.

Inspired by previous works on image amodal segmentation~\cite{ozguroglu2024pix2gestalt, li2016amodal,  follmann2019learning}, we construct our dataset automatically by identifying amodal objects in videos and overlaying occlusions. Yet, naively placing occluders randomly in each frame introduces flickering, compromising consistency. As explained in SAM2~\cite{ravi2024sam}, progressive training on increasingly challenging data enhances model performance. However, it remains unclear what modes of data are more challenging, necessitating a model-in-the-loop approach, where model feedback guides data augmentation. As a result, we finally curate the OvO dataset with two difficulty levels: OvO-Easy and OvO-Hard. The overall curation pipeline, illustrated in \cref{fig:data_pipeline}, consists of three steps.

\noindent\textbf{Identify Candidate} We start by collecting real-world videos from diverse sources, including MVImgNet~\cite{yu2023mvimgnet}, Bdd100k~\cite{yu2020bdd100k}, and SA-V~\cite{ravi2024sam}. Object instances are segmented within videos using either provided annotations~\cite{yu2020bdd100k, ravi2024sam} or off-the-shelf segmentation models~\cite{ren2024grounded}.

\noindent\textbf{Amodal Check} We then identify high-quality and fully visible candidates using provided annotations~\cite{yu2020bdd100k} or a combination of heuristic rules and manual filtering. The heuristic rules are based on three main criteria: (1) Instances surrounded by regions closer to the camera, are likely to be occluded; (2) Tiny segmentations or those touching boundaries imply less prominent objects; and (3) Segmentations with numerous internal holes are likely of low quality. 

\noindent\textbf{Progressive Occlusion Overlay} Once suitable amodal candidates are selected, we apply occlusions by consistently overlaying images of occluders sourced from SA-1B~\cite{kirillov2023segment}. 

We begin by constructing a OvO-Easy dataset with a relatively naive yet consistent occlusion overlay strategy. Occlusions are guaranteed in both the first and last frame, with an occlusion rate between $0.3$ and $0.7$. To create smooth and consistent motion throughout the video, the occluders' positions are linearly interpolated across intermediate frames.

During iterative testing, we identify failure cases where the model struggles, such as regions with severe or persistent occlusions, uncommon or swift object motions, and occlusions caused by close or direct contact. Guided by these observations, we further curate an OvO-Hard dataset with a more aggressive occlusion overlay strategy, featuring a more challenging setting. It first sets the occluder's position in the first frame, maintaining an occlusion rate between $0.4$ and $0.8$. The occluder then moves in tandem with the amodal candidate's bounding box, dynamically adjusting its size according to changes in the candidate's bounding box. This not only results in a higher occlusion rate compared to OvO-Easy but also introduces a ``persistent occlusion'' effect, where a portion of the candidate is likely to be persistently occluded throughout the video, necessitating a stronger completion capability. To further enhance realism, image feathering techniques are applied, softening occlusion boundaries for mimicking occlusions by close contact. 

In addition to progressive occlusion overlays, the choice of data sources plays a crucial role. To boost the diversity of OvO-Hard, we incorporate image-level datasets, which offer a broader range of object instances with precise annotations. Specifically, we select complete object instances from SA-1B~\cite{kirillov2023segment} and apply various image transformation techniques, such as zooming, parallel moving, and warping, to simulate camera movement and object dynamics. These transformed images are then sequenced into synthetic videos, to which we apply the same curation strategy used for video data. An illustration of these image transformation techniques is provided in \supp.

\vspace{-0.6em}
\section{Method}
\vspace{-0.3em}

\begin{figure*}[ht!]
    \centering
    \includegraphics[width=0.95\linewidth]{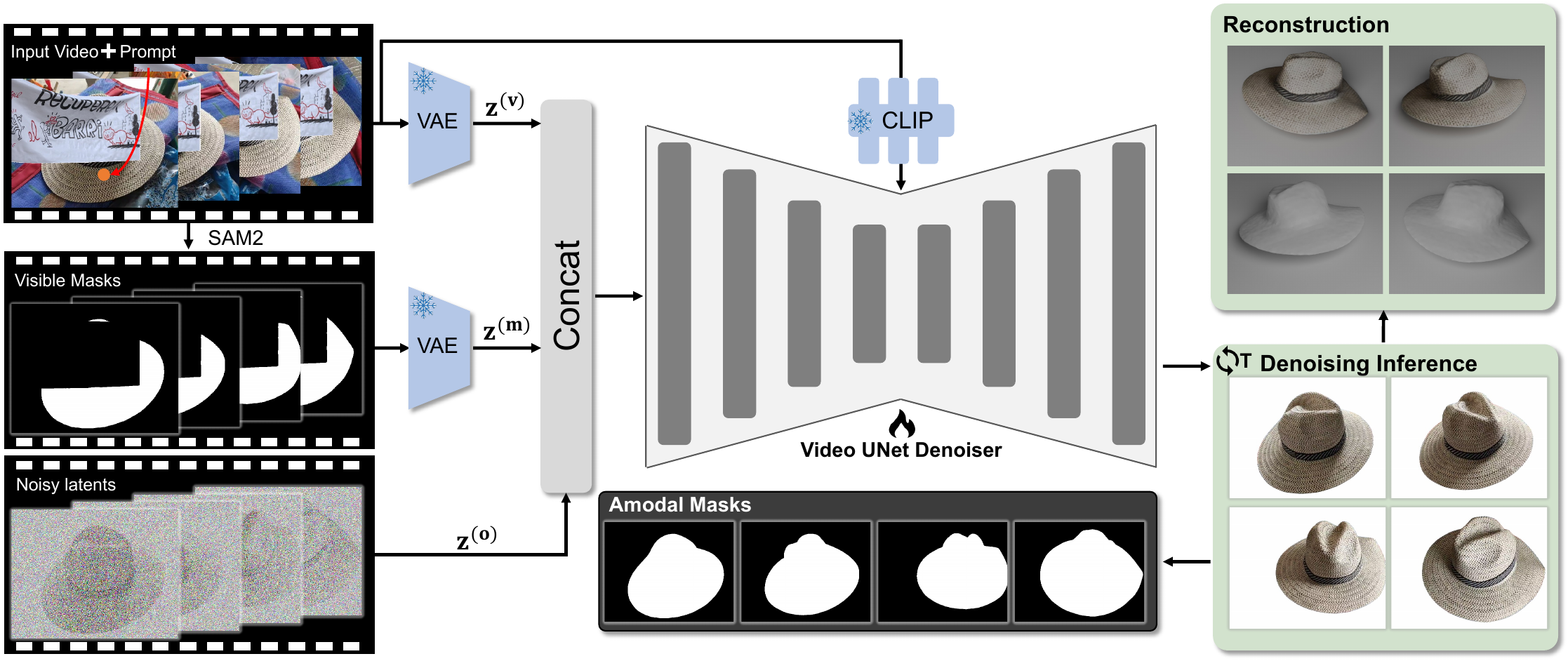}
    \caption{\textbf{Method overview.} We propose \model, a conditional video diffusion model for \acf{vac}. \model takes an occluded video and a visual prompt (point or mask) as input and synthesizes a consistent and complete object throughout the video.}
    \label{fig:method_overview}
    \vspace{-1em}
\end{figure*}

We address the task of \Acf{vac}. Formally, given a video $\mathbf{v} \in \mathbb{R}^{T \times H \times W \times 3}$ and prompt $p$, which is a point or mask identifying the object of interest, the target model $f$ aims to recover a consistent unoccluded video of the object $\mathbf{o} \in \mathbb{R}^{T \times H \times W \times 3}$, \ie,
\begin{equation}
\label{eq0}
\mathbf{o} = f(\mathbf{v}, p).
\end{equation}
Given a prompt $p$, we use SAM2~\cite{ravi2024sam} to obtain visible masks of the object in a video.
Key insights of our model lie in leveraging the pre-trained video diffusion models and progressively fine-tuning them on our scalable dataset to ensure robust generalization to in-the-wild scenarios. 
An illustrative overview is provided in \cref{fig:method_overview}.

\subsection{Preliminaries of Video Diffusion Models}

\label{sec:method_prelim}
Diffusion models ~\cite{ho2020denoising, song2020denoising} learn a target data distribution $p(\mathbf{x})$ by gradually adding noise and then iteratively denoising. In the context of video diffusion, a sample $\mathbf{x}_0 \sim p(\mathbf{x})$ represents a video, and we choose the publicly available Stable Video Diffusion (SVD)~\cite{blattmann2023stable} as our base model in this paper. SVD adopts an EDM-framework~\cite{karras2022elucidating} as the noise scheduler. In the forward process, \iid Gaussian noise with $\sigma_{t}^2$-variance will be added to the data $\mathbf{x}_0 \sim p(\mathbf{x})$:
\begin{equation}\label{eq1}
\mathbf{x}_t = \mathbf{x}_0 + \sigma_{t}^2\epsilon, \quad \epsilon \sim \mathcal{N}(\mathbf{0}, \mathbf{I}).
\end{equation}
$\mathbf{x}_t \sim p(\mathbf{x};\sigma_t)$ stands for the data with noise level $\sigma_t$, and when the noise level is large enough, denoted as $\sigma_{\text{max}}$, the distribution will be indistinguishable from pure Gaussian noise. Building on this fact, SVD starts from a high-variance Gaussian noise $\mathbf{x}_M \sim \mathcal{N}(\mathbf{0}, \sigma_{\text{max}}^2\mathbf{I})$, and gradually denoise it towards $\sigma_0 = 0$ for clean data using a learnable denoiser $D_{\theta}$, which is trained via denoising score matching:
\begin{equation}\label{eq2}
\mathbb{E}_{\mathbf{x}_t \sim p(\mathbf{x};\sigma_{t}), \sigma_t \sim p(\sigma)}[\lambda_{\sigma_t}||D_{\theta}(\mathbf{x}_t; \sigma_t; \textbf{c}) - \mathbf{x}_0||_{2}^{2}],
\end{equation}
where $p(\sigma)$ is the noise level distribution, $\mathbf{c}$ is the conditional information, and $\lambda_{\sigma_t}$ is the loss weight at time $t$.

\subsection{\model}
As shown in \cref{fig:method_overview}, our proposed method builds upon a pre-trained video diffusion model, aiming to consistently complete a target object given an occluded video $\mathbf{v}$ and the object's corresponding visible masks.

\paragraph{Conditioning on an entire video} 
Video diffusion models typically use two types of conditional guidance: (1) High-level conditioning that leverages CLIP~\cite{radford2021learning} embedding of a reference image $y$ via cross-attention in the U-Net, and (2) Low-level conditioning that concatenates VAE~\cite{kingma2013auto} encoded reference images with noisy frames $\mathbf{x}_t^{1:N}$ along the channel dimension as input to the U-Net.

SVD~\cite{blattmann2023stable} conditions on a single reference image. In our \ac{vac} task, however, the entire video serves as input, requiring a conditioning mechanism that spans all frames. Inspired by previous works~\cite{shao2025learning, hu2024depthcrafter, van2024generative} and as illustrated in \cref{fig:method_overview}, we directly concatenate the input video latent $\mathbf{z}^{(\mathbf{v})}$ with the noisy object video latent $\mathbf{z}^{(\mathbf{o})}$. To incorporate high-level semantic information, we inject frame-wise CLIP embeddings into corresponding frames of the denoising U-Net through cross-attention. This design provides the model with a broad receptive field, enabling it to access information from all frames simultaneously, which is crucial for maintaining consistency when decoding amodal objects. 

The visible masks $\mathbf{m} \in \mathbb{R}^{T \times H \times W}$, pinpointing the object of interest and mitigating the risk of hallucinating irrelevant content, are encoded into the same latent space as the input video, denoted as $\mathbf{z}^{(\mathbf{m})}$. These latent masks are concatenated with the video latents $\mathbf{z}^{(\mathbf{v})}$ and the noisy object latents $\mathbf{z}^{(\mathbf{o})}$, forming the final input to the denoising U-Net.

\paragraph{Progressive training strategy}
Inspired by SAM2~\cite{kirillov2023segment} which improves performance by learning from increasingly harder data, we adopt a similar progressive training strategy. In \textit{Phase 1}, we start training from a relatively simple dataset, OvO-Easy, possessing a low occlusion rate and few ``persistent occlusion'' as explained in \cref{sec:method_data}. This dataset draws exclusively from MVImgNet~\cite{yu2023mvimgnet} and SA-V~\cite{ravi2024sam}, facilitating a smooth initial training phase. Through iterative testing, we identify failure cases where the model struggles, such as regions with severe or persistent occlusions, uncommon or swift object motions, and occlusions caused by close or direct contact. In response, we further curate the OvO-Hard dataset and continue \textit{Phase 2} training. This involves adopting a more aggressive occlusion overlay strategy, incorporating more diverse sources from image-level dataset SA-1B~\cite{kirillov2023segment}, and applying image feathering techniques to create more realistic occlusions. To further validate the effectiveness of progressive fine-tuning, we continue training the \textit{Phase 1} and \textit{Phase 2} checkpoints on an autonomous driving dataset Bdd100k~\cite{yu2020bdd100k}, which is curated using the same strategy to curate OvO-Hard but possesses a dramatically different resolution compared to OvO-Easy and OvO-Hard. The ablative results of progressive training are presented in \cref{sec:exp_ablation}. As mentioned by previous works~\cite{li2024puppet, van2024generative}, SVD struggles with substantial resolution changes, so we refrain from joint training on this split.

\begin{table*}[t!]
\small
\centering
\caption{\textbf{Quantitative comparison on amodal completion and segmentation.} We evaluate on four synthetic datasets, and our method outperforms baselines in terms of image alignment, mask alignment, and cross-frame consistency.}
\begin{tabular}{ccccccccc}
\toprule
\multirow{2}[2]{*}{Dataset} & \multirow{2}[2]{*}{Method} & \multicolumn{3}{c}{Image Alignment} & \multicolumn{1}{c}{Mask Alignment} & \multicolumn{2}{c}{Frame Consistency} \\
\cmidrule(lr){3-5} \cmidrule(lr){6-6} \cmidrule(lr){7-8}
 &  & PSNR($\uparrow$) & SSIM($\uparrow$) & LPIPS($\downarrow$) & IoU($\uparrow$) & CLIP-T($\uparrow$) & FVD($\downarrow$)\\
\midrule
\multirow{4}{*}{OvO-Easy} 
          & Pix2Video & 12.590 & 0.682 & 0.368 & 42.1 & 0.937 & 1561.10 \\
          & $\mathrm{E}^2\mathrm{FGVI}$ & 13.385 & 0.664 & 0.306 & 59.8 & 0.964 & 1015.17 \\
          & pix2gestalt & 20.699 & 0.829 & 0.109 & 88.9 & 0.926 & 389.77 \\
          & Ours & \textbf{26.110} & \textbf{0.906} & \textbf{0.054} & \textbf{95.6} & \textbf{0.976} & \textbf{142.37} \\
\midrule
\multirow{4}{*}{OvO-Hard}
          & Pix2Video & 11.873 & 0.662 & 0.402 & 35.2 & 0.945 & 1715.87 \\
          & $\mathrm{E}^2\mathrm{FGVI}$ & 12.861 & 0.639 & 0.346 & 59.5 & 0.967 & 1260.60 \\
          & pix2gestalt & 18.184 & 0.777 & 0.157 & 83.7 & 0.913 & 674.24 \\
          & Ours & \textbf{22.573} & \textbf{0.857} & \textbf{0.093} & \textbf{92.4} & \textbf{0.979} & 
          \textbf{291.49}  \\
\midrule
\multirow{6}{*}{Kubric-Static}
          & Pix2Video & 14.181 & 0.680 & 0.338 & 34.1 & 0.936 & 1166.68 \\
          & $\mathrm{E}^2\mathrm{FGVI}$ & 14.194 & 0.684 & 0.230 & 57.5 & 0.964 & 738.05 \\
          & pix2gestalt & 20.516 & 0.822 & 0.125 & 75.6 & 0.915 & 382.67 \\
          & PCNet & 19.848 & 0.857 & 0.118 & 83.2 & 0.959 & 254.08 \\
          & Diffusion-VAS & 21.358 & 0.845 & 0.101 & \textbf{84.3} & \textbf{0.971} & 228.05 \\
          & Ours & \textbf{23.963} & \textbf{0.891} & \textbf{0.073} & 83.9 & 0.970 & \textbf{162.91} \\
\midrule
\multirow{6}{*}{Kubric-Dynamic}
          & Pix2Video & 15.571 & 0.728 & 0.311 & 20.8 & 0.942 & 979.57 \\
          & $\mathrm{E}^2\mathrm{FGVI}$ & 14.617 & 0.691 & 0.224 & 50.6 & 0.969 & 726.55 \\
          & pix2gestalt & 19.676 & 0.830 & 0.132 & 68.3 & 0.906 & 429.11 \\
          & PCNet & 20.373 & 0.864 & 0.104 & 76.2 & 0.957 & 253.45 \\
          & Diffusion-VAS & 21.067 & 0.859 & 0.096 & \textbf{77.8} & \textbf{0.977} & 230.02 \\
          & Ours & \textbf{23.054} & \textbf{0.886} & \textbf{0.080} & 77.4 & 0.970 & \textbf{209.28} \\
\bottomrule
\end{tabular}
\vspace{-1em}
\label{tab:completion}
\end{table*}
\paragraph{Tackling long videos}
Current video generative models~\cite{xing2025dynamicrafter, blattmann2023stable, hong2022cogvideo}, including ours, are constrained by a fixed frame number. To overcome this, we introduce a sliding window mechanism that incrementally synthesizes future frames based on previously generated outputs. To be specific, given a video clip $\mathbf{v}^{1:N}$, where $N$ is the total frame number, our model begins with the initial $\mathbf{v}^{1:k}\ (k = 14)$ frames, producing a completed object video $\mathbf{o}^{1:k}$. The synthesized object is then reintegrated into the video as:
\begin{equation}\label{eq4}
\tilde{\mathbf{v}}^{1:k} = \mathbf{o}^{1:k} \odot \mathcal{M} + \mathbf{v}^{1:k} \odot (1 - \mathcal{M}),
\end{equation}
where $\mathcal{M}$ denotes the object mask area obtained from $\mathbf{o}^{1:k}$ and $\odot$ represents pixel-wise product. In the next iteration of completion, our model takes as input the concatenation of previously generated frames and the next sequence of frames, denoted as $[\tilde{\mathbf{v}}^{(k - m + 1):k};\mathbf{v}^{(k + 1):(2k - m)}]$, where $m = 5$ represents the sliding window size. This process iterates until the entire video clip is processed. By reinserting the synthesized objects into subsequent frames, the model preserves better consistency across long video sequences.

\label{sec:method_model}

\section{Experiment}
\begin{figure*}[t]
    \centering
    \includegraphics[width=0.95\linewidth]{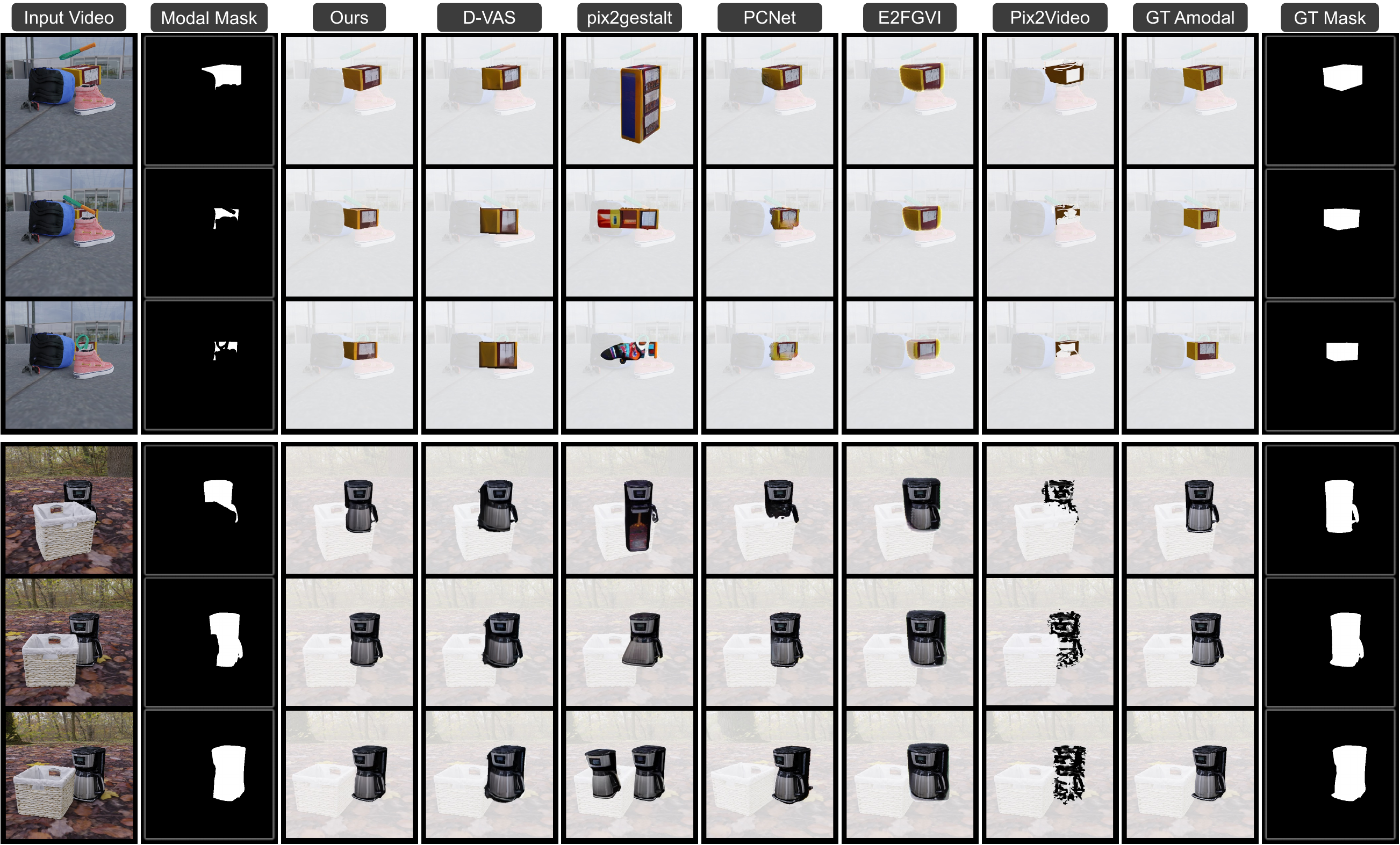}
    \caption{\textbf{Qualitative comparison on Kubric~\cite{greff2022kubric}}. The first example is from Kubric-Dynamic and the second from Kubric-Static. Our method achieves better consistency and can effectively extract information from nearby frames to handle highly occluded scenarios.}
    \label{fig:completion_quali}
    \vspace{-1em}
\end{figure*}

\subsection{Video Amodal Completion and Segmentation}
\label{sec:exp_complete}
\paragraph{Setup} \acf{vas} predict the amodal masks of occluded objects consistently across video frames, while \acf{vac} requires further filling in reasonable content. We quantitatively benchmark VAC on four datasets: OvO-Easy, OvO-Hard, Kubric-Static, Kubric-Dynamic ~\cite{greff2022kubric, downs2022google}, and VAS on two datasets: MOVi-B and MOVi-D~\cite{gao2023coarse}. We also conduct a user study on 20 in-the-wild~\cite{ebert2021bridge, walke2023bridgedata, dai2017scannet, yeshwanth2023scannet++, yu2020bdd100k, xu2018youtube} and Internet videos. Details are in \supp. 
\paragraph{Baselines and Metrics} We compare our method against pix2gestalt~\cite{ozguroglu2024pix2gestalt} and PCNet~\cite{zhan2020self} for image-level amodal completion, a video inpainting method $\mathrm{E}^2\mathrm{FGVI}$~\cite{li2022towards}, a video editing approach Pix2Video~\cite{ceylan2023pix2video}, as well as a video amodal method Diffusion-VAS~\cite{chen2024using}. Since PCNet requires identifying all objects within a scene, we are only able to evaluate it on Kubric datasets. Amodal segmentation masks for all methods are obtained by thresholding their predicted frames. For VAS, we also compare against EoRaS~\cite{fan2023rethinking}, with its pre-trained checkpoints on MOVi-B and MOVi-D.
We utilize PSNR, SSIM~\cite{wang2004image}, and LPIPS~\cite{zhang2018unreasonable} to evaluate image alignment with ground-truth reference frames, mIoU for amodal mask alignment, and CLIP-T~\cite{esser2023structure} and FVD~\cite{unterthiner2018towards} to assess cross-frame consistency.
Further details are provided in \supp.

\begin{table}[H]
        \begin{minipage}[t]{0.52\linewidth}
            \centering
            \vspace{-1em}
            \captionof{table}{\textbf{User Study}.} 
            \vspace{-1.0em}
            \resizebox{\linewidth}{!}{
                \begin{tabular}{lcccc}
                    \toprule
                    Method & CQ ($\uparrow$) & CC ($\uparrow$) & CP ($\uparrow$)\\
                    \midrule
                    Pix2Video & 2.20 & 2.44 & 2.13  \\
                    pix2gestalt & 2.88 & 2.44 & 2.63  \\
                    $\mathrm{E}^2\mathrm{FGVI}$ & 2.60 & 2.95 & 2.53 \\
                    Ours & \textbf{4.65} & \textbf{4.63} & \textbf{4.64} \\
                    \bottomrule
                \end{tabular}
            }
            \label{tab:user}
        \end{minipage}
        \begin{minipage}[t]{0.46\linewidth}
            \centering
            \vspace{-1em}
            \captionof{table}{\textbf{VAS results}.}
            \vspace{-1em}
            \resizebox{\linewidth}{!}{
                \begin{tabular}{lccc}
                \toprule
                Method & Zero-shot & MOVi-B & MOVi-D \\
                \midrule
                $\mathrm{E}^2\mathrm{FGVI}$ & $\checkmark$& 41.9 & 31.2  \\
                P2G(Top-1) & $\checkmark$ & 56.4 & 48.3  \\ 
                P2G(Top-3) & $\checkmark$& 57.2 & 48.9 \\ 
                Ours(Top-1) & $\checkmark$& 73.1 & 56.6  \\ 
                Ours(Top-3) & $\checkmark$& \textbf{77.8} & 61.3  \\ 
                EoRaS & $\times$& 75.8 & \textbf{69.4}  \\ 
                \bottomrule
                \end{tabular}
            }
            \label{tab:vas}
        \end{minipage}
        \vspace{-1em}
\end{table}
\paragraph{Results} We present \ac{vac} comparison results in \cref{tab:completion} and qualitative examples in \cref{fig:completion_quali}, with user study in \cref{tab:user}. VAS comparison results on MOVi-B and MOVi-D are in \cref{tab:vas}. Our method outperforms baselines in the task of content completion, achieving significantly better performance in terms of Content Quality (CQ), Content Consistency (CC), and Content Plausibility (CP), as validated by user studies involving 36 participants and \cref{tab:completion}. However, Diffusion-VAS achieves slightly better segmentation performance, likely due to its explicit prediction of amodal masks. On \ac{vas} benchmarks, our method surpasses zero-shot baselines. From visualized results, $\mathrm{E}^2\mathrm{FGVI}$~\cite{li2022towards} and Pix2Video~\cite{ceylan2023pix2video} fail to generate reasonable amodal content. PCNet~\cite{zhan2020self} and pix2gestalt~\cite{ozguroglu2024pix2gestalt} generate plausible content in a single frame but struggle with cross-frame consistency and fail in highly occluded scenarios, as shown in the first example in \cref{fig:completion_quali}. Diffusion-VAS is capable of generating plausible content across frames, but the results often exhibit noticeable artifacts and visual glitches.

\begin{figure}[h!]
    \centering
    \includegraphics[width=1.0\linewidth]{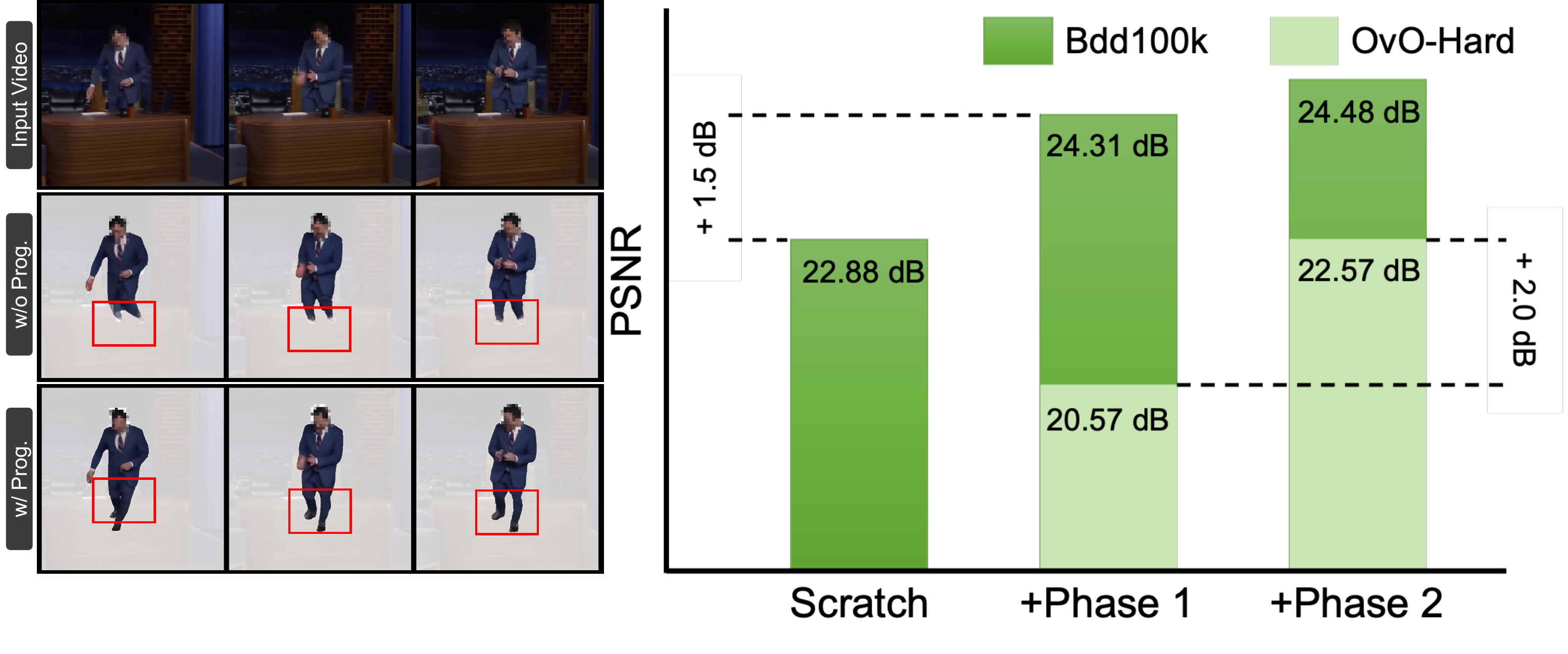}
    \vspace{-1.5em}
    \caption{\textbf{Ablation study.} We test our model's performance \textbf{w/} and \textbf{w/o} progressive training on OvO-Hard and Bdd100k test set.}
    \label{fig:ablation}
    \vspace{-1.5em}
\end{figure}

\begin{figure*}[h]
    \centering
    \includegraphics[width=0.95\linewidth]{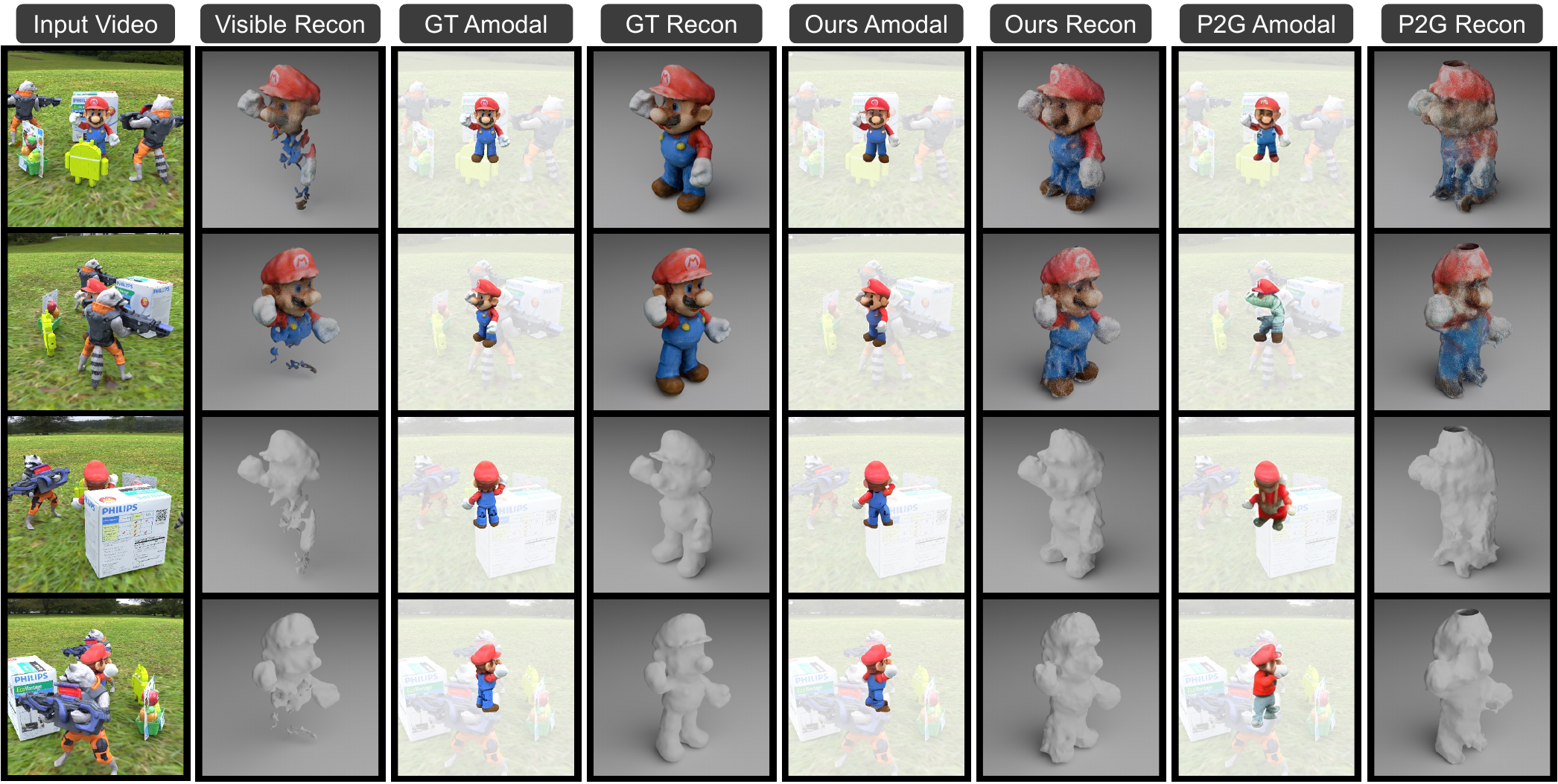}
    \caption{\textbf{Visualized comparison on reconstruction}. Our method is capable of reconstructing more complete and higher-quality textured mesh, further demonstrating the consistency of our synthesized results. `P2G' denotes pix2gestalt~\cite{ozguroglu2024pix2gestalt}.}
    \label{fig:recon_quali}
    \vspace{-0.5em}
\end{figure*}

\subsection{Ablation Study}
\label{sec:exp_ablation}
To assess the effectiveness of our progressive training paradigm, we conduct an ablation study in \cref{fig:ablation}. The qualitative results show that models trained progressively synthesize more complete and coherent content. Additionally, we quantitatively benchmark performance on Bdd100k and OvO-Hard in \cref{fig:ablation}. The \textit{Phase 1} and \textit{Phase 2} correspond to the progressive training stages in \cref{sec:method_model}. For Bdd100k, \textit{Scratch} refers to training without progressive learning, inheriting weights directly from SVD. In contrast, \textit{Phase 1} and \textit{Phase 2} significantly benefit its performance by building upon weights from corresponding phases on OvO.

\subsection{Object Reconstruction}
\label{sec:exp_recon}
\paragraph{Setup} We demonstrate that \ac{vac} significantly enhances object reconstruction, particularly in challenging occluded scenarios. We benchmark our approach on a Compositional-GSO dataset~\cite{downs2022google} with 20 scenes, leveraging NeRF2Mesh~\cite{tang2023delicate} to reconstruct objects from synthesized images. Further details are in \supp.


\paragraph{Results} We compare 3D reconstruction results using only the visible parts of objects against those generated by our model, pix2gestalt~\cite{ozguroglu2024pix2gestalt}, and $\mathrm{E}^2\mathrm{FGVI}$~\cite{li2022towards}. For reference, we also include reconstructions using GT amodal images. We utilize Chamfer Distance (CD), F-Score, and Normal Consistency (NC) as metrics, with quantitative results in \cref{tab:recon_met} and qualitative comparison in \cref{fig:recon_quali}. Our method showcases significantly better reconstruction quality, closely aligning with GT. The `Visible' baseline lacks information in occluded areas, leading to significant missing parts of the object during reconstruction. While pix2gestalt~\cite{ozguroglu2024pix2gestalt} can synthesize content in occluded regions, it fails to guarantee multi-view consistency, leading to compromised reconstructions. In contrast, our method effectively synthesizes consistent multi-view images, providing a solid foundation for high-quality object reconstruction.
\begin{table}
\centering
\small
\caption{\textbf{Quantitative comparison on object reconstruction}. Our method provides consistent amodal objects that lead to better geometry compared with baselines.}
\vspace{-1mm}
\begin{tabular}{lccc}
\toprule
Method & CD ($\downarrow$) & F-Score ($\uparrow$) & NC ($\uparrow$)\\
\midrule
$\mathrm{E}^2\mathrm{FGVI}$ & 7.57 & 22.34 & 56.16 \\
Visible & 4.39 & 34.86 & 65.99 \\ 
pix2gestalt & 1.97 & 45.30 & 75.04 \\ 
Ours & \textbf{1.26} & \textbf{57.71} & \textbf{81.09} \\ 
\midrule
GT Amodal & 1.05 & 66.01 & 84.41 \\
\bottomrule
\end{tabular}
\label{tab:recon_met}
\vspace{-2mm}
\end{table} 

\subsection{6-DoF Object Pose Estimation}

\begin{table}
\centering
\small
\caption{\textbf{Quantitative comparison on pose estimation}. Our method predicts plausible content in occluded areas, offering additional information that improves the accuracy of pose estimation.}
\resizebox{\linewidth}{!}{
\begin{tabular}{lcccc}
\toprule
Method & Rot-Acc ($\uparrow$) & trans-Acc ($\uparrow$) & Rot-Err ($\downarrow$) & trans-Err ($\downarrow$)\\
\midrule
w/o ours & 26.5 & 11.3 & 41.5 & 67.8 \\ 
w/ ours & \textbf{34.7} & \textbf{13.8} & \textbf{39.1} & \textbf{59.1} \\
\bottomrule
\end{tabular}
}
\label{tab:pose_met}
\end{table}
\paragraph{Setup} We test if video amodal completion can improve 6-DoF object pose estimation in occluded scenarios. Given a video, we first synthesize the complete object and integrate it back into the original scene. We then compare the performance of pose estimation with and without our method, using an off-the-shelf pose estimation module POPE~\cite{fan2024pope}.

\begin{figure}[h]
    \centering
    \includegraphics[width=0.8\linewidth]{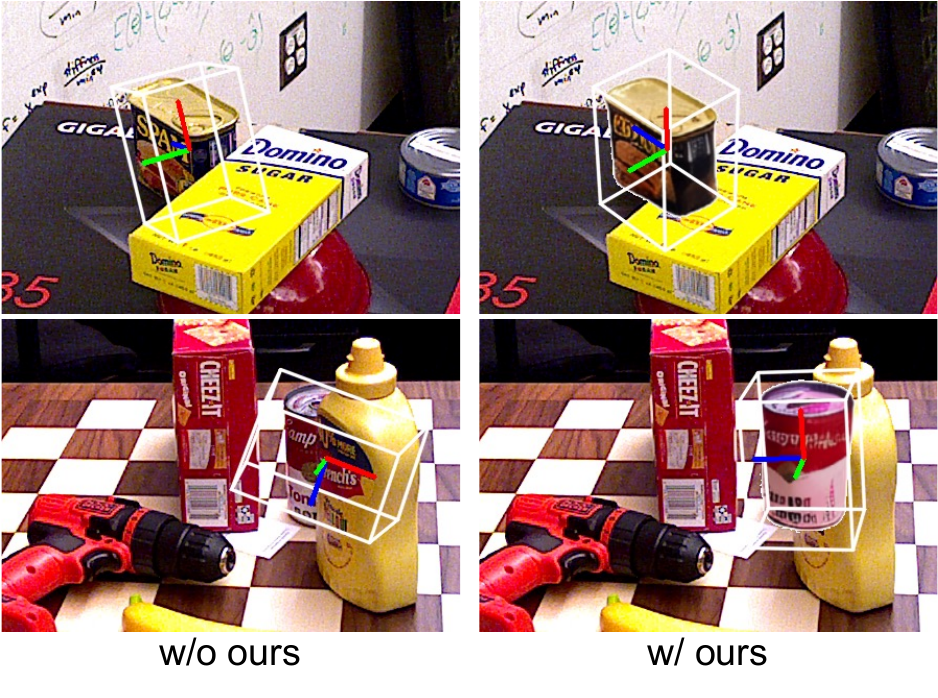}
    \vspace{-2.5mm}
    \caption{\textbf{6-DoF pose estimation on YCB-Video~\cite{xiang2017posecnn}.} \model aids accurate pose estimation by synthesizing occluded contents.}
    \label{fig:pose_quali}
    \vspace{-2.5mm}
\end{figure}

\paragraph{Results} We test on a total of 336 image pairs from YCB-Video~\cite{xiang2017posecnn}. From \cref{tab:pose_met}, incorporating our amodal prediction helps POPE to better comprehend the location, orientation, and size of occluded objects. From visualized results in \cref{fig:pose_quali}, POPE predicts more accurate pose, especially orientation, with the help of amodal contents.
\label{sec:exp_detect}

\vspace{-0.5em}
\section{Conclusion}
We propose \model for \acf{vac}, leveraging the rich consistent manifolds learned by pre-trained diffusion models. For robust generalization to in-the-wild videos, we curate a scalable dataset from diverse real-world videos with multiple difficulty levels and introduce a progressive fine-tuning paradigm. We demonstrate \model's generalization capability to diverse in-the-wild videos as well as its potential to aid downstream tasks.
\section*{Acknowledgment}
This work is supported by the Sichuan Science and Technology Program (2023YFSY0008), China Tower-Peking University Joint Laboratory of Intelligent Society and Space Governance, National Natural Science Foundation of China (61632003, 61375022, 61403005), Grant SCITLAB20017 of Intelligent Terminal Key Laboratory of SiChuan Province, Beijing Advanced Innovation Center for Intelligent Robots and Systems (2018IRS11), and PEK-SenseTime Joint Laboratory of Machine Vision. This work is also supported by the State Key Laboratory of General Artificial Intelligence.

\small
\bibliographystyle{ieeenat_fullname}
\bibliography{main}

\clearpage
\appendix
\renewcommand{\thefigure}{S.\arabic{figure}}
\renewcommand{\thetable}{S.\arabic{table}}
\renewcommand{\theequation}{S.\arabic{equation}}
\maketitlesupplementary

\noindent We provide details on data curation, implementation, additional qualitative results, and a discussion of limitations in \supp. 
\section{Data Curation Details}
In this section, we elaborate on the details of how we curate the \acf{ovo} dataset, including how to overlay occluders consistently throughout the video in \cref{data:overlay}, how we apply image transformation techniques to augment our dataset with image-level datasets in \cref{data:image}, and other curation details in \cref{data:details}.
\subsection{Overlay Occluders Consistently}
\label{data:overlay}
\begin{figure}[ht]
    \centering
    \includegraphics[width=\linewidth]{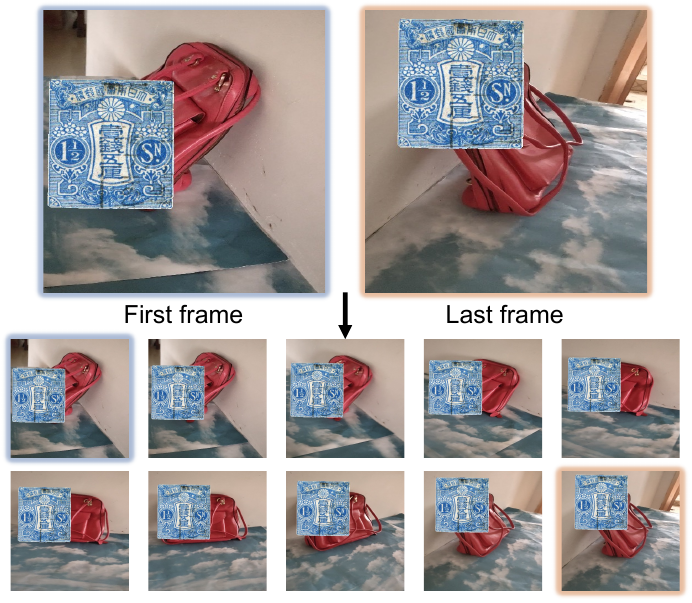}
    \caption{\textbf{Illustration on consistent occlusion.} In \acs{ovo}-Easy, we first select a proper position and scale in the first and last frame, and then interpolate in intermediate frames. The stamp is the occluder in the example. }
    \label{fig:occ_illus}
\end{figure}

To ensure consistent occlusions, it is crucial to maintain continuous change in the occluders' properties across video frames. This includes the occluders' position, $\mathbf{p}$, which specifies their position, and scale, $\mathbf{s}$, which determines their size. We employ two heuristic strategies in \acs{ovo}-Easy and \acs{ovo}-Hard dataset to generate occlusions. 
\paragraph{\acs{ovo}-Easy} Occluders are selected from Objaverse~\cite{deitke2023objaverse, deitke2023objaversexl} and SA-1B~\cite{kirillov2023segment}. Appropriate occlusion positions are identified for the first and last frames of the video. The occlusion rate, defined as the ratio of the occluded area to the total area of the object, is constrained to lie between $0.3$ and $0.7$ in both the first and last frames. For the intermediate frame $i$ , the occlusion position $\mathbf{p}_{i}$ and scale $\mathbf{s}_{i}$ are determined through linear interpolation, ensuring smooth transitions:
\begin{equation}\label{eq1}
    \mathbf{p}_i = \frac{i}{N} \cdot (\mathbf{p}_{\text{ed}} - \mathbf{p}_{\text{st}}) + \mathbf{p}_{\text{st}}, \
    \mathbf{s}_i = \frac{i}{N} \cdot (\mathbf{s}_{\text{ed}} - \mathbf{s}_{\text{st}}) + \mathbf{s}_{\text{st}},
\end{equation}
where $\mathbf{p}_{\text{st}}, \mathbf{s}_{\text{st}}, \mathbf{p}_{\text{ed}}, \mathbf{s}_{\text{ed}}$ denotes the occlusion position and scale in the first and last frame, and $N$ stands for the total frame number.
An example is illustrated in \cref{fig:occ_illus}. Apart from a relatively low occlusion rate, some objects are fully visible in intermediate frames.

\paragraph{\acs{ovo}-Hard} In contrast, \acs{ovo}-Hard begins by selecting an initial occlusion position $\mathbf{p}_\text{st}$ and scale $\mathbf{s}_\text{st}$ in the first frame. The position and size of the occluder are then dynamically adjusted throughout the video, guided by changes in the bounding box of the occluded object:
\begin{equation}\label{eq2}
    \mathbf{p}_i = \mathbf{c}_i - \mathbf{c}_\text{st} + \mathbf{p}_\text{st}, \
    \mathbf{s}_i = max(\frac{\mathbf{h}_i}{\mathbf{h}_\text{st}}, \frac{\mathbf{w}_i}{\mathbf{w}_\text{st}})^{1/3} \cdot \textbf{s}_\text{st},
\end{equation}
where $\mathbf{c}_\text{st}$ and $\mathbf{c}_i$ stands for the center of the bounding box in the initial frame and frame $i$, $\mathbf{h}_i$, $\mathbf{w}_i$, $\mathbf{h}_\text{st}$, and $\mathbf{w}_\text{st}$ stands for the height and width of the $i-th$ frame and the initial frame. In \acs{ovo}-Hard, we also apply image feathering techniques to blend the occluder more naturally into the image. The occlusion rate of the initial frame is constrained to lie between $0.4$ and $0.8$. 
\subsection{Image Transformations}
\label{data:image}
\begin{figure}[ht]
    \centering
    \includegraphics[width=\linewidth]{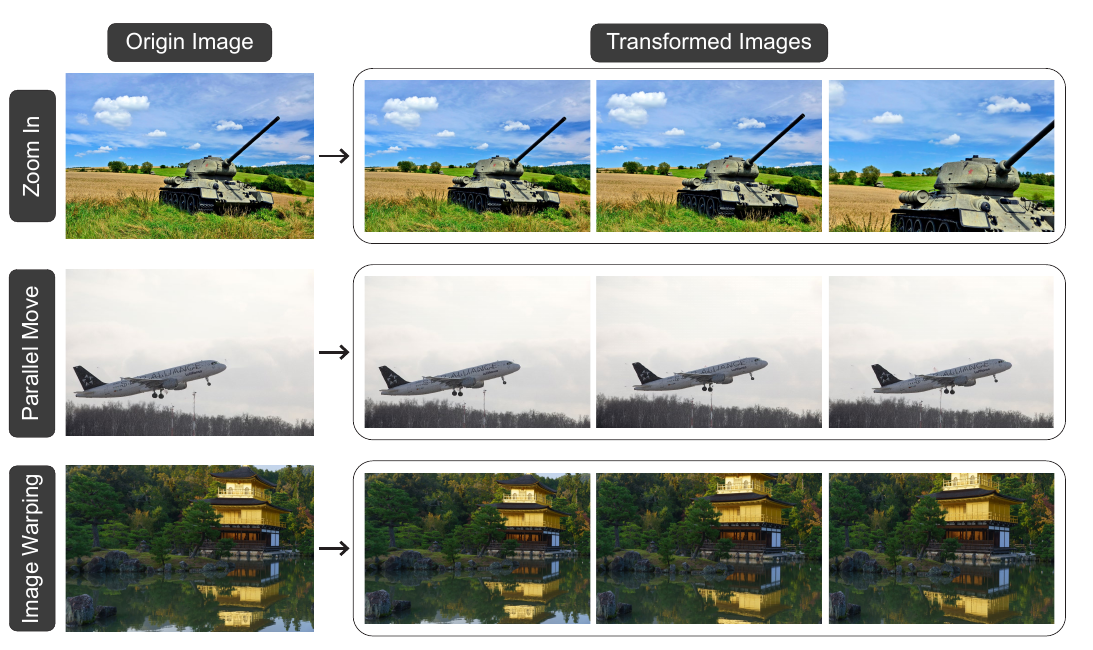}
    \caption{\textbf{Image transformations.} We augment our \ac{ovo} dataset by incorporating image-level datasets and image transformation techniques to simulate videos.}
    \label{fig:image_trans}
\end{figure}
In addition to introducing more severe occlusions, we further augment our dataset in \acs{ovo}-Hard using the image-level dataset SA-1B~\cite{kirillov2023segment}. Specifically, we apply three image transformation techniques: zooming, parallel moving, and image warping. An illustration of these techniques is shown in \cref{fig:image_trans}. The zooming transformation is implemented by center-cropping the image, while image warping is achieved through a homography transformation. For parallel moving, we first segment the complete foreground object using the provided annotations~\cite{ozguroglu2024pix2gestalt, kirillov2023segment}. The background area is then inpainted, after which the foreground and background are shifted in parallel to create a motion effect. For instance, in \cref{fig:image_trans}, the plane visually appears to move to the right.
\subsection{Other Details}
\label{sec:dataset}
In this section, we will elaborate on the data sources, the process of occluder selection, and the details of amodal check including heuristic rules and manual filtering.
\paragraph{Data sources}
To ensure the diversity of our dataset \acs{ovo}, we conducted experiments using subsets of four datasets. After applying the amodal check, the final dataset consists of approximately 90K videos from MVImgNet~\cite{yu2023mvimgnet}, 11K videos from SA-V~\cite{ravi2024sam}, 10K videos from Bdd100k~\cite{yu2020bdd100k}, and 24K videos from SA-1B~\cite{kirillov2023segment}. For constructing \acs{ovo}-Easy, we use data from MVImgNet~\cite{yu2023mvimgnet} and SA-V~\cite{ravi2024sam}. For the more challenging \acs{ovo}-Hard, we further add data from Bdd100k~\cite{yu2020bdd100k} and SA-1B~\cite{kirillov2023segment}. Additionally, we reserve approximately 1K videos as the test sets for \acs{ovo}-Easy and \acs{ovo}-Hard, which are used to benchmark our method against various baselines.
\paragraph{Occluders}
\begin{figure}[ht]
    \centering
    \includegraphics[width=\linewidth]{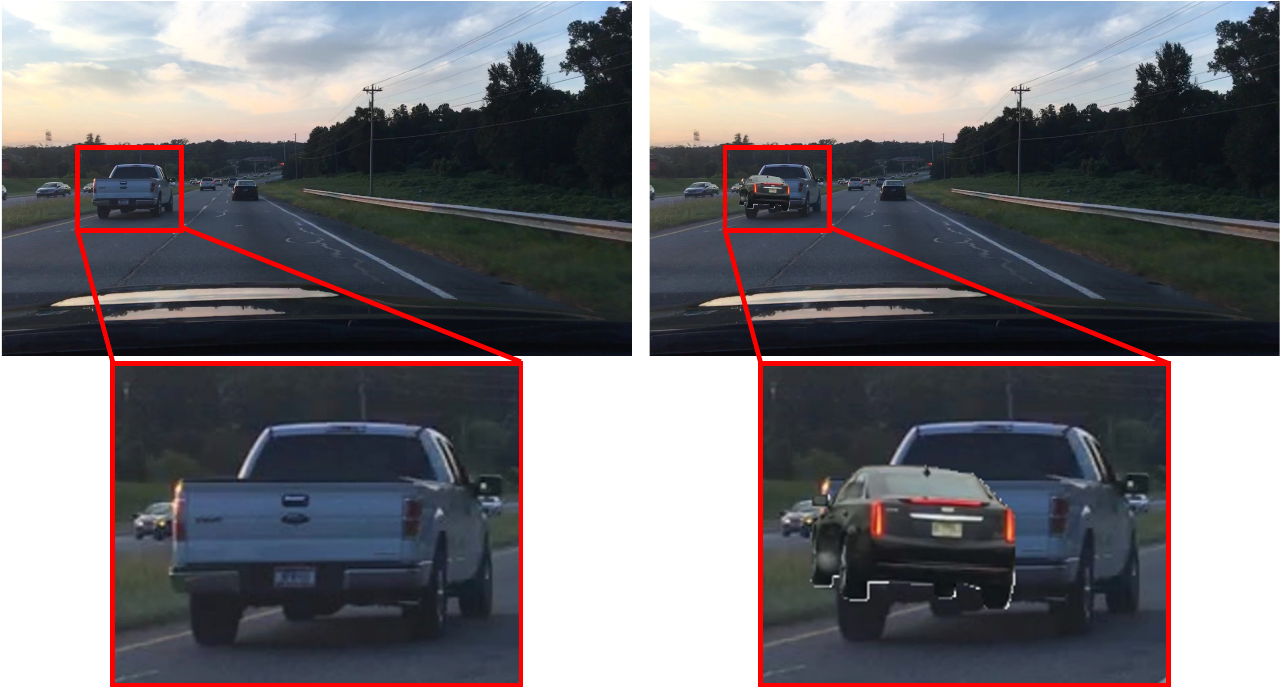}
    \caption{\textbf{Illustration on Bdd100K occluders.} We select occluders from Bdd100K since many of the occluders in SA-1B are not applicable to the autonomous driving context.}
    \label{fig:bdd100k}
\end{figure}
In \acs{ovo}-Easy, we select approximately 50K occluders from a subset of SA-1B~\cite{kirillov2023segment} and around 20K occluders from a subset of Objaverse~\cite{deitke2023objaverse}. For occluders sourced from SA-1B, objects are segmented using the provided annotations, ensuring that the occluder occupies a significant portion of the image. For occluders from Objaverse, we render rotation videos consisting of 40 frames using Blender~\cite{blender}.
In \acs{ovo}-Hard, occluders from Objaverse are excluded to enhance realism. Additionally, when curating data pairs for Bdd100k~\cite{yu2020bdd100k}, we select occluders directly from Bdd100k~\cite{yu2020bdd100k} to maintain domain relevance, as many occluders from SA-1B and Objaverse are not applicable to the autonomous driving context. An example on the occluders of Bdd100k is shown in \cref{fig:bdd100k}. 
\paragraph{Amodal check}
\begin{figure}[ht]
    \centering
    \includegraphics[width=\linewidth]{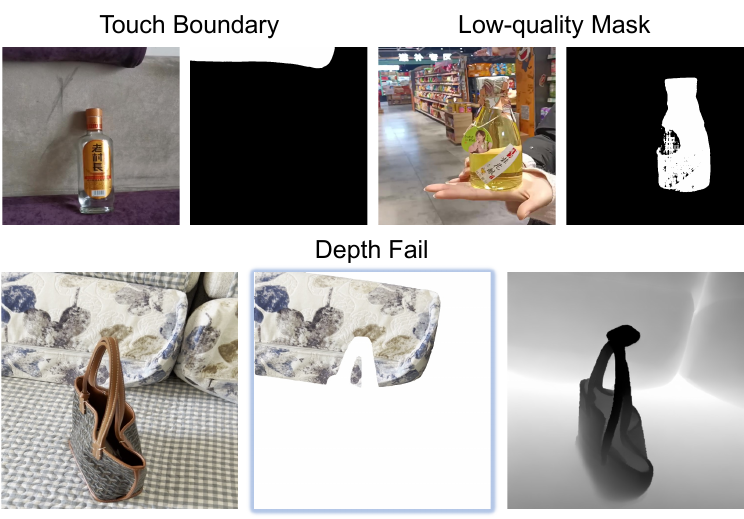}
    \caption{\textbf{Illustration on the heuristic amodal check.} We apply three heuristic rules to check whether an object is complete. }
    \label{fig:fail_illus}
\end{figure}
We apply three heuristic rules to filter out candidates likely to be incomplete, as illustrated in \cref{fig:fail_illus}. In the first example, the mask touches the image boundary, indicating potential incompleteness. The second example shows a mask with numerous internal holes, suggesting an inaccurate segmentation. In the third example, the pillow is positioned behind the foreground bag, failing the depth consistency check. Despite these heuristic rules, some incomplete object candidates remain. To address this, we leverage crowdsourcing to further filter out incomplete samples that have incorrectly passed the heuristic amodal check.
\label{data:details}
\section{Implementation Details}
In this section, we provide detailed explanations of the training and inference process in \cref{sec:training}, the dataset, metrics and baselines used for amodal completion and segmentation in \cref{exp:completion}, the dataset curation and reconstruction pipeline for object reconstruction in \cref{exp:reconstruction}, the intermediate results for pose estimation in \cref{exp:pose}, and details on user study in \cref{exp:user_study}.
\subsection{Training and Inference Details}
\label{sec:training}
We trained our model on the \acs{ovo}-Easy dataset for 7 epochs and continued training on the \acs{ovo}-Hard dataset for another 7 epochs, resulting in a total training time of approximately 6 days using 8 NVIDIA A800 (80G) GPUs. Due to computational constraints, all input and target videos were resized to a resolution of $384 \times 384$. The batch size per GPU was set to 4, yielding a total batch size of 32. To balance the dataset, data pairs from SA-V~\cite{ravi2024sam} were sampled twice per epoch. As noted in prior work~\cite{li2024puppet, van2024generative}, Stable Video Diffusion (SVD)\cite{blattmann2023stable} is not robust to variations in resolution. To address this, we further fine-tuned the model on the Bdd100k subset for 8 epochs at a resolution of $640 \times 384$, taking around 20 hours. This is because the typical resolution used in autonomous driving scenarios significantly differs from those in MVImgNet\cite{yu2023mvimgnet} and SA-V~\cite{ravi2024sam}. 
We employed the SVD version configured to predict 14 frames. To handle additional visible mask inputs, extra channels in the first layer were added after concatenation and initialized to zero. A freeze motion bucket and fixed frame rate were used for simplicity. During inference, conditional samples were generated using the EDM sampler with 50 steps~\cite{karras2022elucidating}, taking approximately 20 seconds to produce an output video.
\subsection{Amodal Completion and Segmentation}
\label{exp:completion}
\paragraph{Test Dataset}
In addition to the test split of \acs{ovo}-Easy and \acs{ovo}-Hard, we curate two additional datasets, Kubric-Static and Kubric-Dynamic, using the Kubric simulator~\cite{greff2022kubric} to benchmark the generalizability of the models.
For Kubric-Static, we randomly select 2 to 5 objects from GSO~\cite{downs2022google}, along with a background dome. The objects are placed in a static scene with randomized scales and positions. A rotating video is rendered using blender~\cite{blender} by rotating the camera around the scene, capturing each object's modal and amodal masks as well as amodal RGB images. From the rendered videos, we filter 365 samples with occlusions for benchmarking, ensuring that at least 10 out of 14 frames in each video contain occluded objects.
For Kubric-Dynamic, we randomly select 2 to 3 falling objects and combine them with 2 to 4 static objects placed on the ground as well as a background dome. A linearly changing camera trajectory is applied before rendering each frame. Using the same filtering criteria as in Kubric-Static, we select 323 videos with occlusions for benchmarking.
\paragraph{Metrics} Since a significant portion of the synthesized frames is white, this can inflate the PSNR, SSIM, and LPIPS values. To address this, we crop the synthesized amodal object using a dilated bounding box of the ground-truth (GT) amodal mask and compute the PSNR, SSIM, LPIPS, and CLIP-T metrics within this specific region.
\paragraph{Baselines} 
\begin{figure}[ht!]
    \centering
    \includegraphics[width=\linewidth]{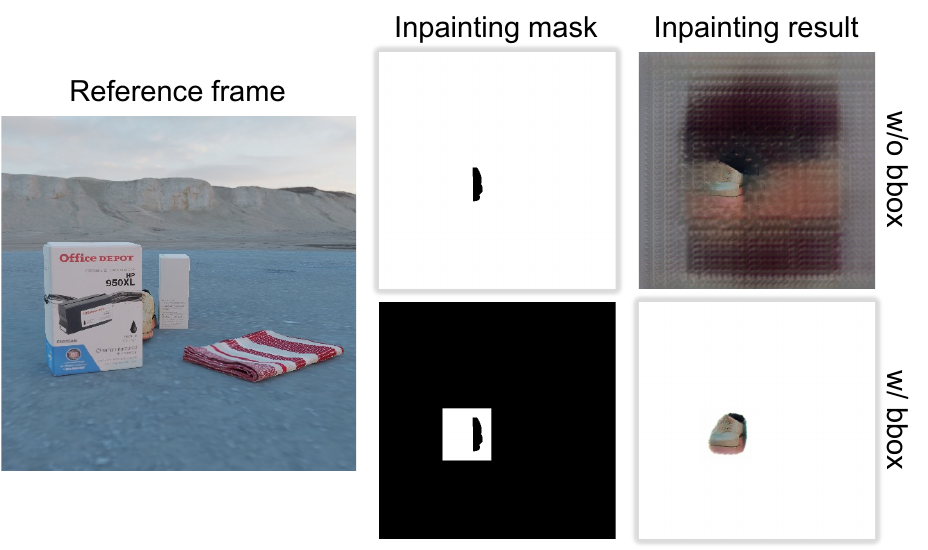}
    \caption{\textbf{Comparison on inpainting mask.} We try two kind of inpainting mask for the $\mathrm{E}^2\mathrm{FGVI}$ baseline, and the one with bounding box constraint is significantly better.}
    \label{fig:baseline_ab}
\end{figure}
Since pix2gestalt~\cite{ozguroglu2024pix2gestalt} can only generate images at a resolution of $256 \times 256$, we resize all results to this resolution when calculating the metrics. For the $\mathrm{E}^2\mathrm{FGVI}$ baseline, we observed significantly degraded performance when masking out all but the visible object area. To address this, we first change the background to white and then use the same dilated bounding box of the ground-truth (GT) amodal mask as the inpainting mask. An illustrative comparison of the inpainting masks and the corresponding results is provided in \cref{fig:baseline_ab}.
\subsection{Object Reconstruction}
\paragraph{Dataset}
We also use objects from GSO~\cite{downs2022google} to create compositional occluded scenes. Specifically, we first place a primary object at the center of the scene. Then, we select 3 to 6 additional objects along with a background dome and arrange them around the central object to ensure occlusion occurs. Finally, we rotate the camera around the center of the scene to generate an occluded video. A total of 20 scenes are composed for benchmarking.
\paragraph{Reconstruction}
\label{exp:reconstruction}
We directly apply NeRF2Mesh~\cite{tang2023delicate} to the synthesized results for reconstruction, where images with better cross-view consistency result in higher reconstruction quality. To enhance performance, NeRF2Mesh also requires an object mask during reconstruction. We obtain this mask by thresholding the synthesized images.
\begin{figure}[ht]
    \centering
    \includegraphics[width=\linewidth]{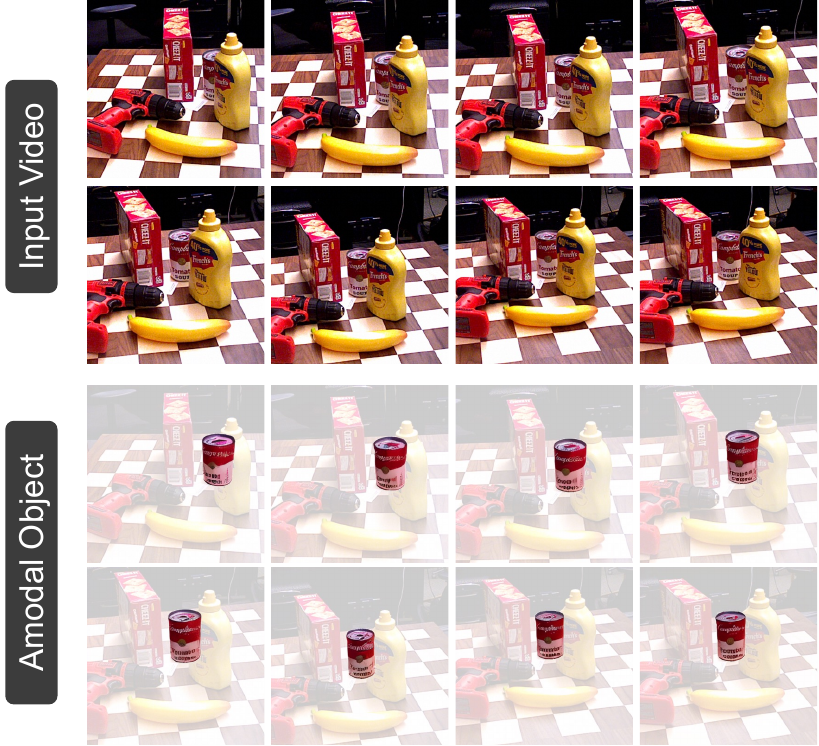}
    \caption{\textbf{Intermediate results on YCB-Video.} We visualize the intermediate synthesized video on YCB-Video.}
    \label{fig:pose_example}
\end{figure}
\begin{figure}[h]
    \centering
    \includegraphics[width=\linewidth]{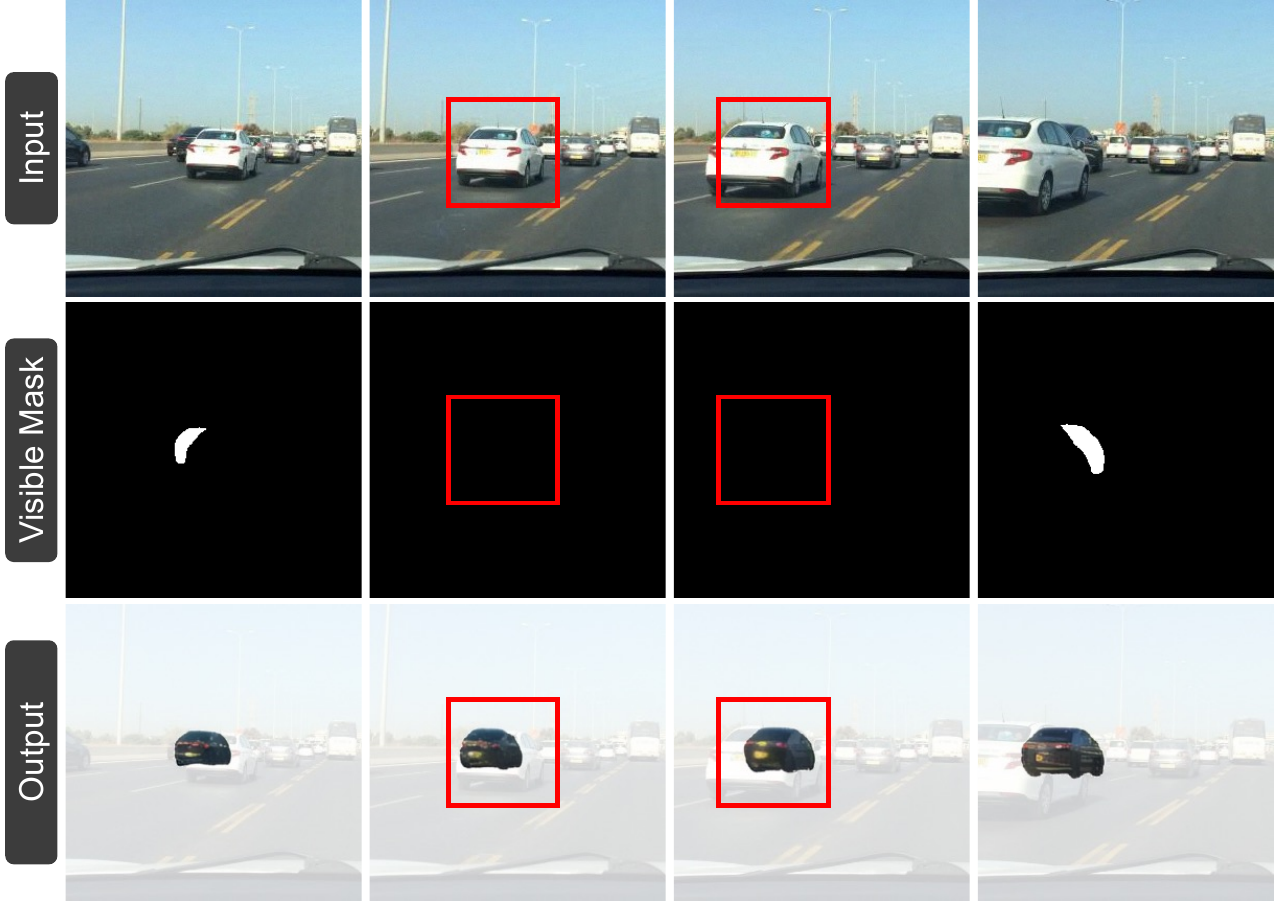}
    \caption{\textbf{Revealing missing objects.} Our method is able to reveal missing objects by inferring from neighboring frames. }
    \label{fig:missing}
\end{figure}
\begin{figure}[h]
    \centering
    \includegraphics[width=\linewidth]{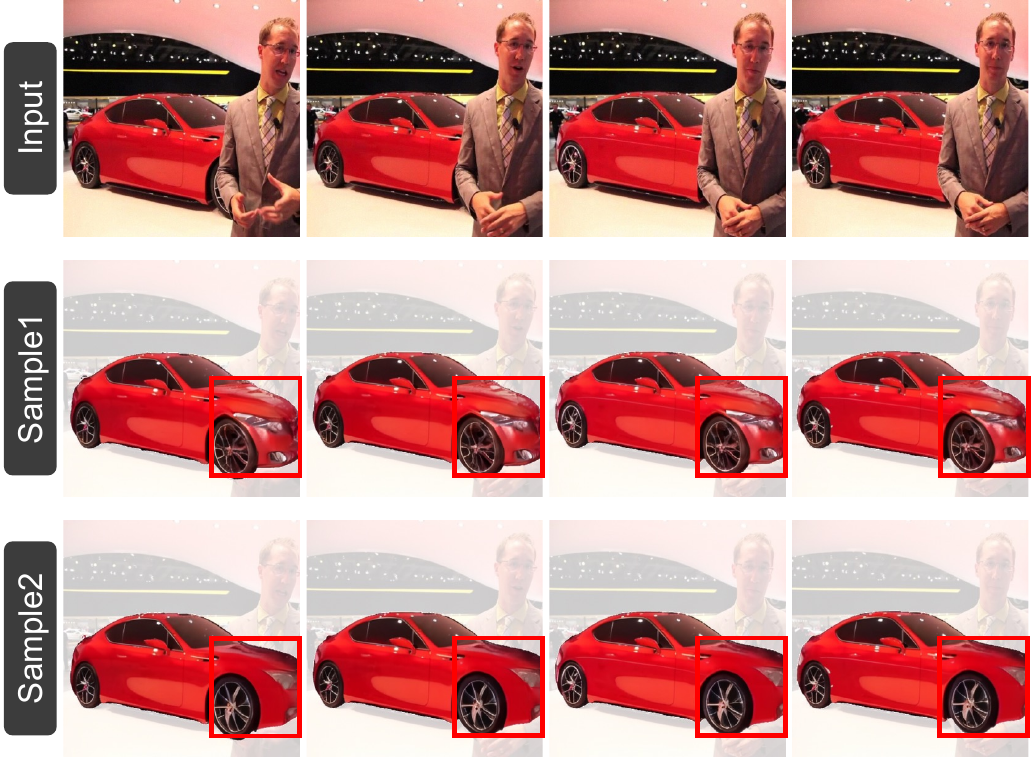}
    \caption{\textbf{Diversity in sampling.} We can sample multiple reasonable results due to the inherent ambiguity in occluded area.}
    \label{fig:diversity}
\end{figure}
\begin{figure}[h]
    \centering
    \includegraphics[width=\linewidth]{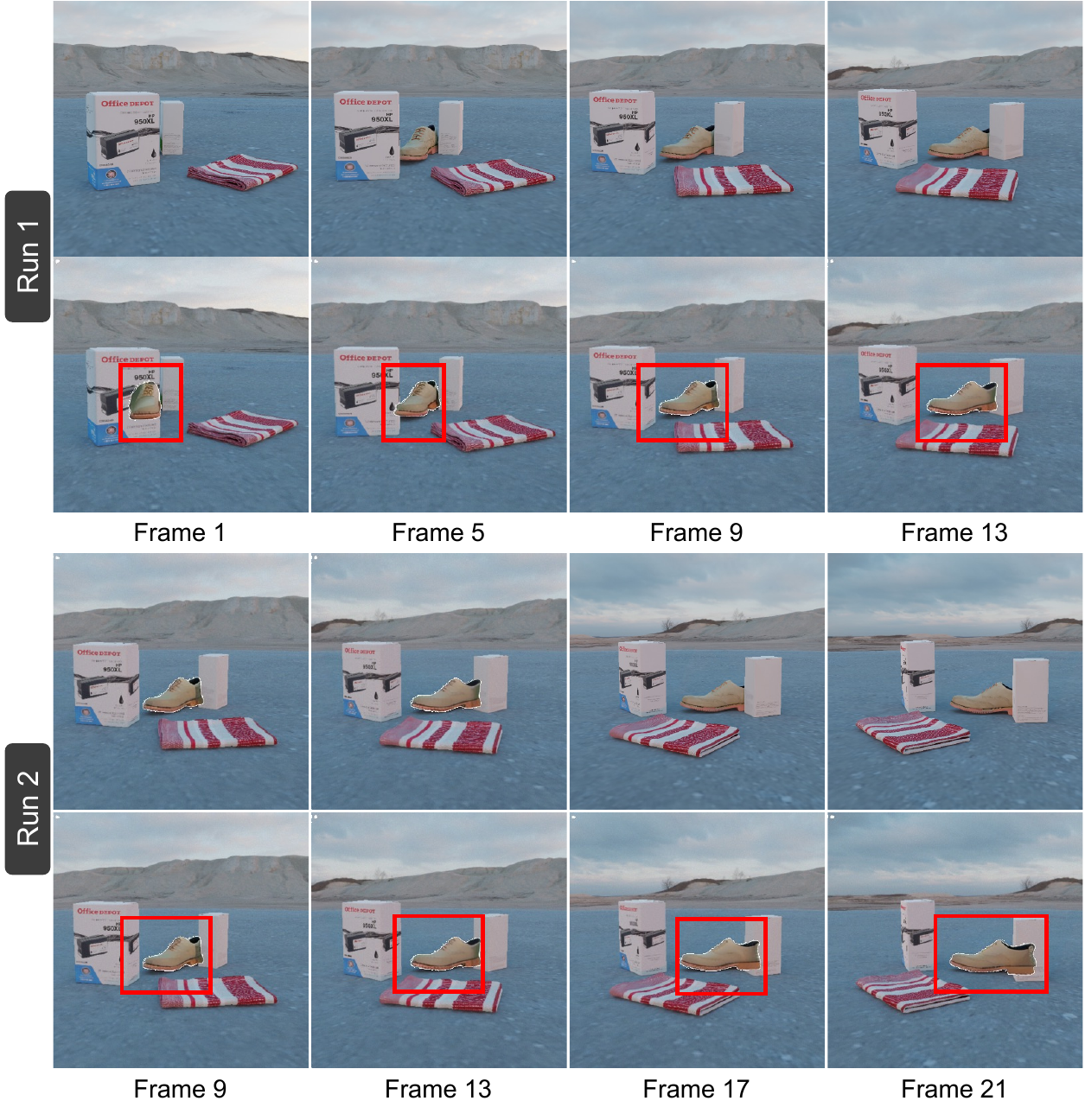}
    \caption{\textbf{Tackling long videos.} Our method is able to extend beyond 14 frames by progressive generation.}
    \label{fig:long}
\end{figure}
\begin{figure}[h]
    \centering
    \includegraphics[width=\linewidth]{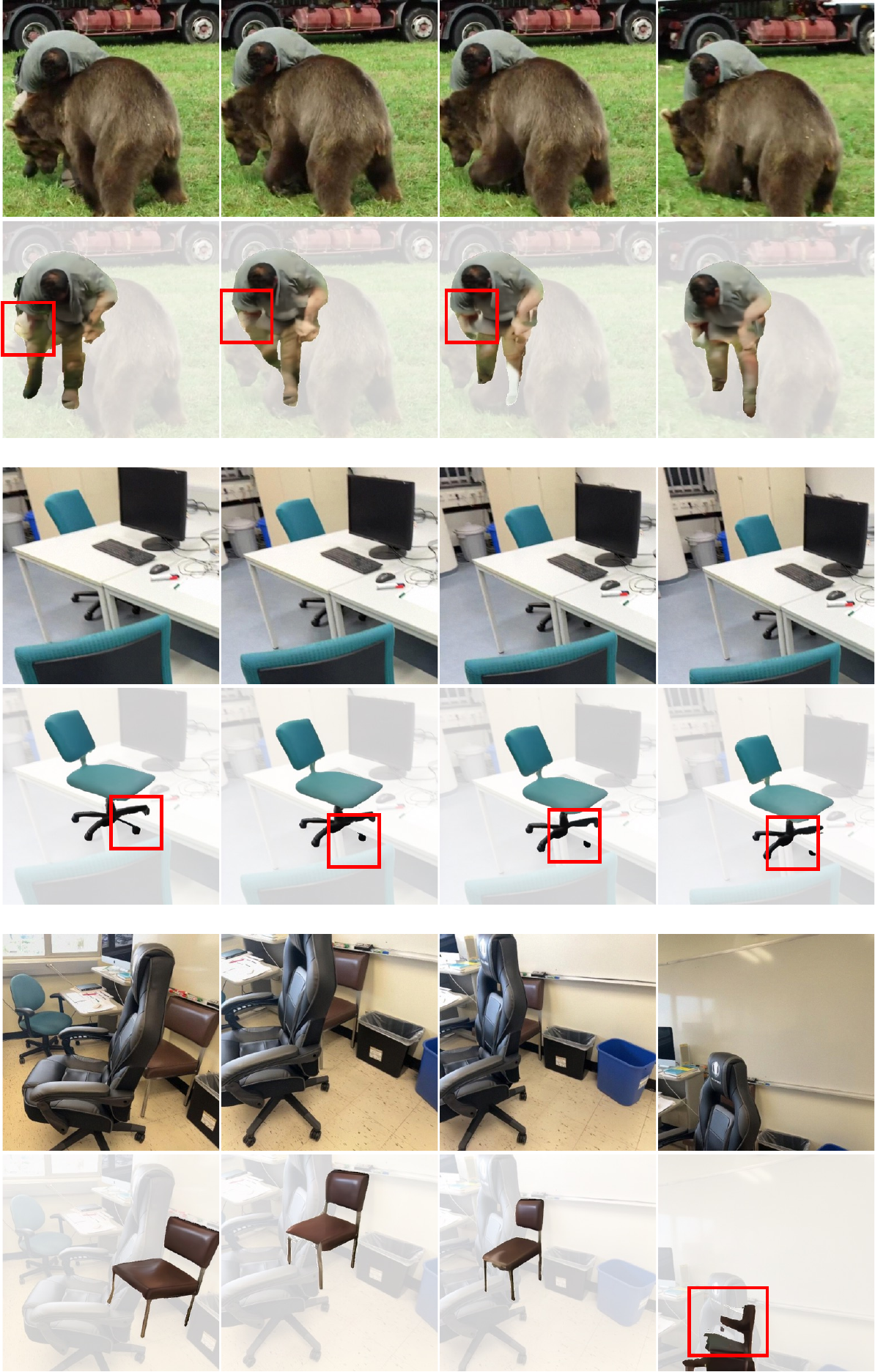}
    \caption{\textbf{Failure cases.} Our method has limitations under certain challenging conditions. For instance, it may produce blurry results in areas with complex occlusions, as seen in the hand region of the first example. Similarly, it struggles to handle extremely fine-grained and heavily occluded structures, as demonstrated in the leg area of the second example. Additionally, our method may fail to perform effectively during drastic and sudden camera movements as shown in the last frame in the third example.}
    \vspace{-1em}
    \label{fig:fail}
\end{figure}
\begin{figure}[h]
    \centering
    \includegraphics[width=\linewidth]{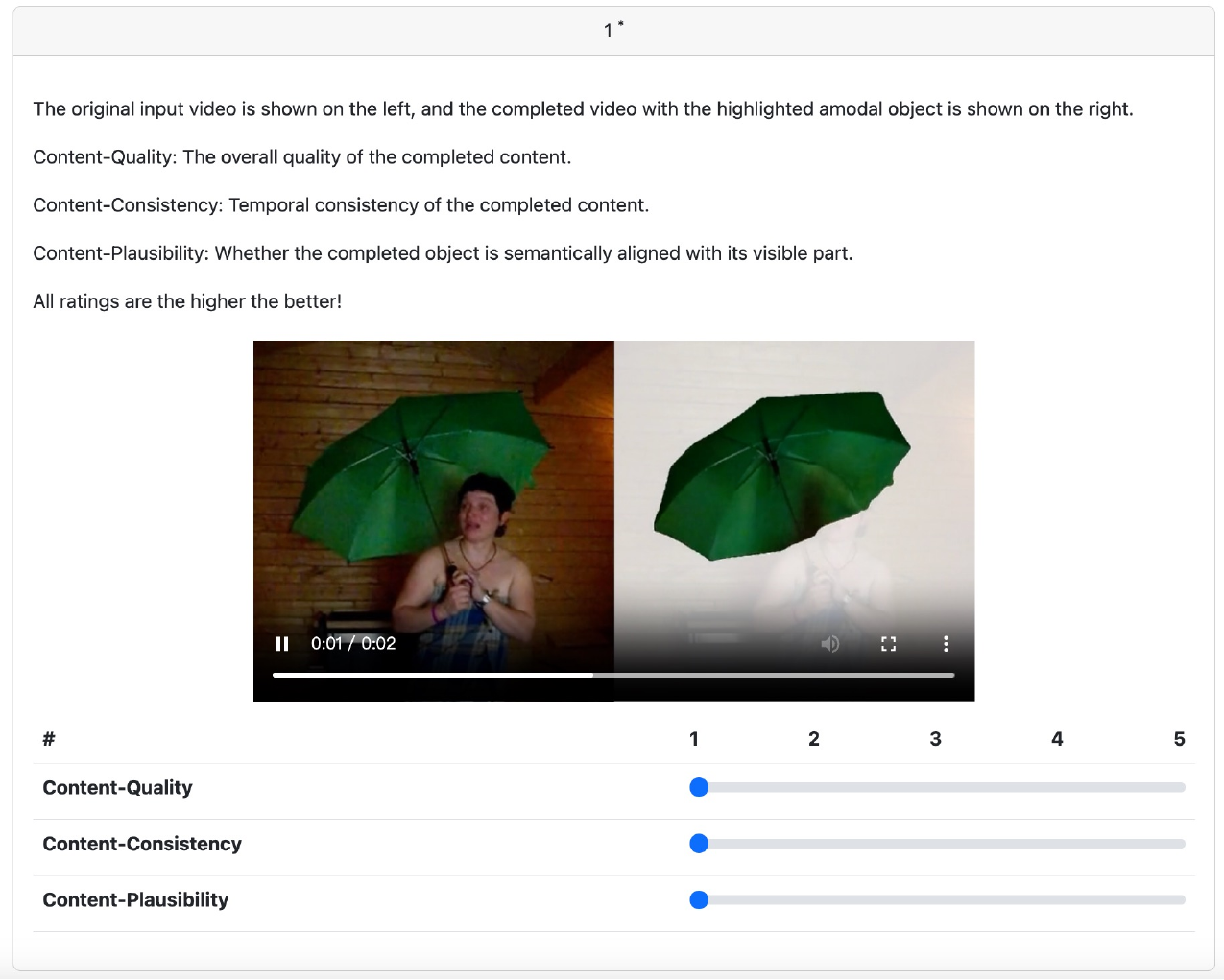}
    \caption{\textbf{Interface of the user study.}}
    \vspace{-1em}
    \label{fig:interface}
\end{figure}
\subsection{Pose Estimation}
\label{exp:pose}
We select a video in YCB-Video~\cite{xiang2017posecnn} and utilize SAM2~\cite{ravi2024sam} to acquire the visible masks of an object throughout the video. An intermediate result (synthesized video of the amodal object) using our method is shown in \cref{fig:pose_example}.

\subsection{User Study}
\label{exp:user_study}
We select 20 videos from unseen datasets including ScanNet++~\cite{dai2017scannet, yeshwanth2023scannet++}, BridgeData~\cite{ebert2021bridge, walke2023bridgedata}, YouTube-VOS~\cite{xu2018youtube}, YCB-Video~\cite{xiang2017posecnn}, and various Internet videos. Each questionnaire contains 16 questions, asking participants to evaluate the completion results across three dimensions: 1) Content-Quality: The overall quality of the completed content, 2) Content-Consistency: Temporal consistency of the completed object, and 3) Content-Plausibility: Whether the completed object is semantically aligned with its visible part. A snapshot of the questionnaire interface is shown in \cref{fig:interface}. We gathered feedback from 36 participants, resulting in 576 valid responses.

\section{Additional Results}
We highly recommend browsing the website, which contains comparisons with baselines, more qualitative results on diverse datasets, and examples of long videos. 
\subsection{Image Amodal Completion}
We also evaluate our method on two image-based datasets to test whether our model still works for image-level tasks. We replicate each image $14$ times to simulate static videos.
\begin{itemize}
    \item \textbf{BSDS-A}:  We evaluate amodal segmentation on BSDS-A~\cite{martin2001database}, using the same test split as \textit{pix2gestalt} (P2G). Since each image has multiple annotations and the specific annotations used in P2G are unavailable, we randomly select one annotation per image, resulting in 730 objects across 200 images. We perform single-shot inference for both methods, which may cause discrepancies from the 16-shot setting in P2G paper. Nonetheless, the overall conclusion remains consistent.
    \item \textbf{Kubric-Img}: We test on 750 images from the Kubric-Static dataset with occlusion rates between $30\%$ and $70\%$ to evaluate both amodal completion and segmentation.
\end{itemize}
\begin{table}[h]
    \begin{minipage}[t]{0.5\linewidth}
        \centering
        \captionof{table}{BSDS-A}
        \resizebox{\linewidth}{!}{
            \begin{tabular}{lccc}
                \toprule
                 & Modal & P2G & Ours  \\
                \midrule
                IoU & 57.7 & \textbf{68.9} & 67.6 \\
                \bottomrule
            \end{tabular}
        }
        \label{tab:reb_seg}
    \end{minipage}
    \begin{minipage}[t]{0.48\linewidth}
        \centering
        \captionof{table}{Kubric-Img}
        \resizebox{\linewidth}{!}{
            \begin{tabular}{lcccc}
            \toprule
             & PSNR & SSIM & LPIPS & IoU \\
            \midrule
            P2G & \textbf{17.578} & 0.781 & 0.153 & \textbf{75.6} \\ 
            Ours & 17.471 & \textbf{0.782} & \textbf{0.146} & 74.4 \\ 
            \bottomrule
            \end{tabular}
        }
        \label{tab:reb_comp}
    \end{minipage}
\end{table}

\begin{figure}[h]
    \centering
    \includegraphics[width=\linewidth]{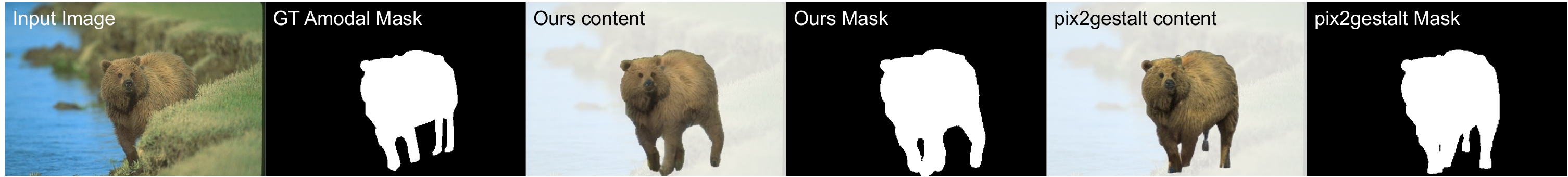}
    \caption{Visualization on BSDS-A}
    \label{fig:reb_img}
\end{figure}
Quantitative results in \cref{tab:reb_seg,tab:reb_comp} as well as visualized results in \cref{fig:reb_img} show our method performs on par with image-based methods, indicating its ability to synthesize plausible amodal content, \textit{even without video context}. We attribute this to the OvO-Hard dataset design, where persistent occlusion forces the model to learn genuinely consistent amodal representations rather than merely rely on video cues. Ensuring such consistency makes VAC innately more complex than image-based tasks.
\subsection{More Comparisons}
\begin{figure*}[t]
    \centering
    \includegraphics[width=\linewidth]{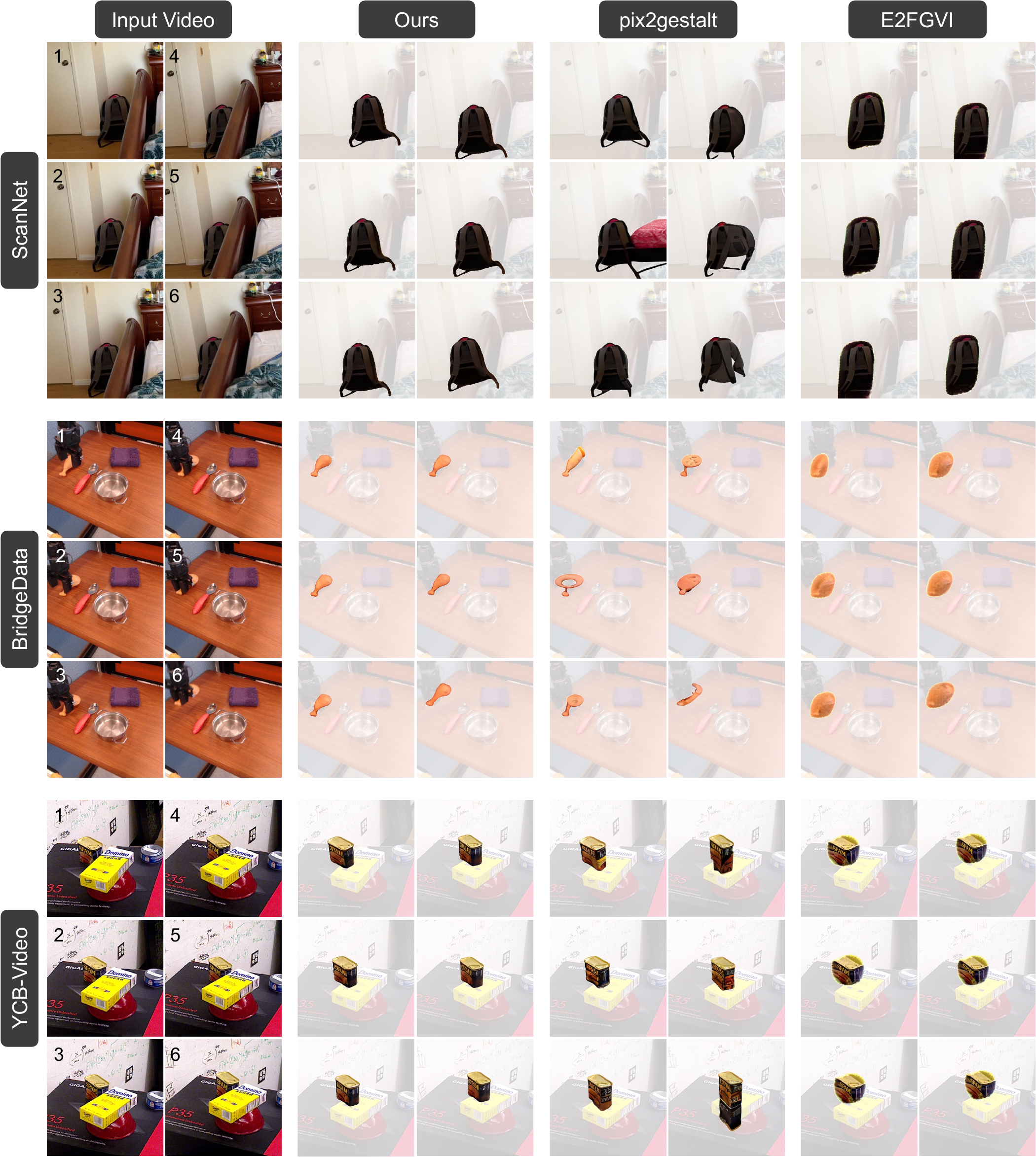}
    \caption{\textbf{Qualitative comparison on ScanNet~\cite{dai2017scannet, yeshwanth2023scannet++}, BridgeData~\cite{ebert2021bridge, walke2023bridgedata}, and YCB-Video~\cite{xiang2017posecnn}.}}
    \label{fig:quali_com_1}
\end{figure*}

\begin{figure*}[t]
    \centering
    \includegraphics[width=\linewidth]{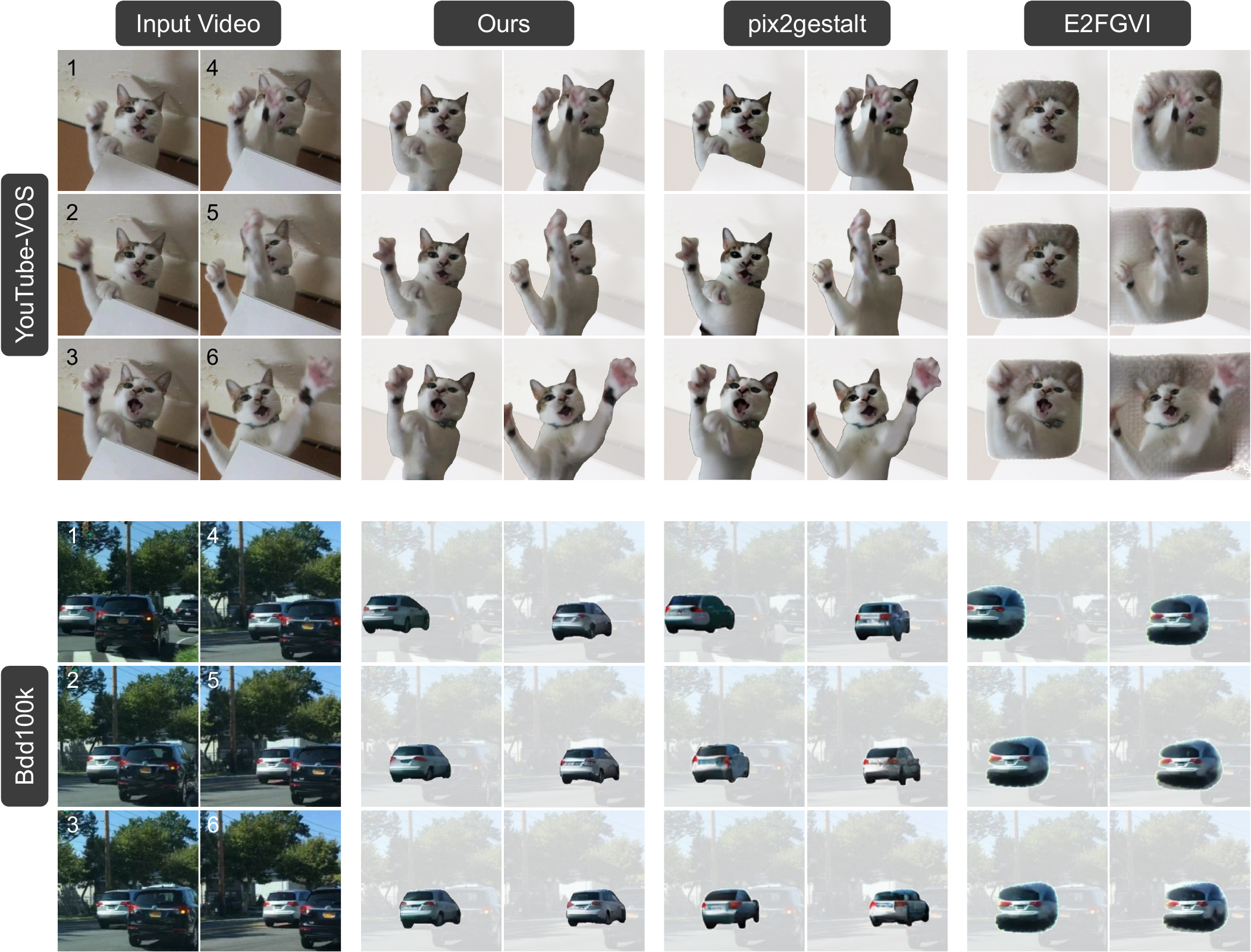}
    \caption{\textbf{Qualitative comparison on YouTube-VOS~\cite{xu2018youtube} and Bdd100k~\cite{yu2020bdd100k}.}}
    \label{fig:quali_com_2}
\end{figure*}

We provide qualitative comparison on ScanNet~\cite{dai2017scannet, yeshwanth2023scannet++}, BridgeData~\cite{ebert2021bridge, walke2023bridgedata}, and YCB-Video~\cite{xiang2017posecnn} in \cref{fig:quali_com_1}. More visualized comparisons on Bdd100k~\cite{yu2020bdd100k} and YouTube-VOS are provided in \cref{fig:quali_com_2}. The original resolution of Bdd100k~\cite{yu2020bdd100k} used for inference is $640 \times 384$, we crop out the region of interest for better visualization. Similarly, for pix2gestalt~\cite{ozguroglu2024pix2gestalt}, we crop a square area as input, as the method requires square images.
\subsection{More Qualitative Results}
\begin{figure*}[t]
    \centering
    \includegraphics[width=\linewidth]{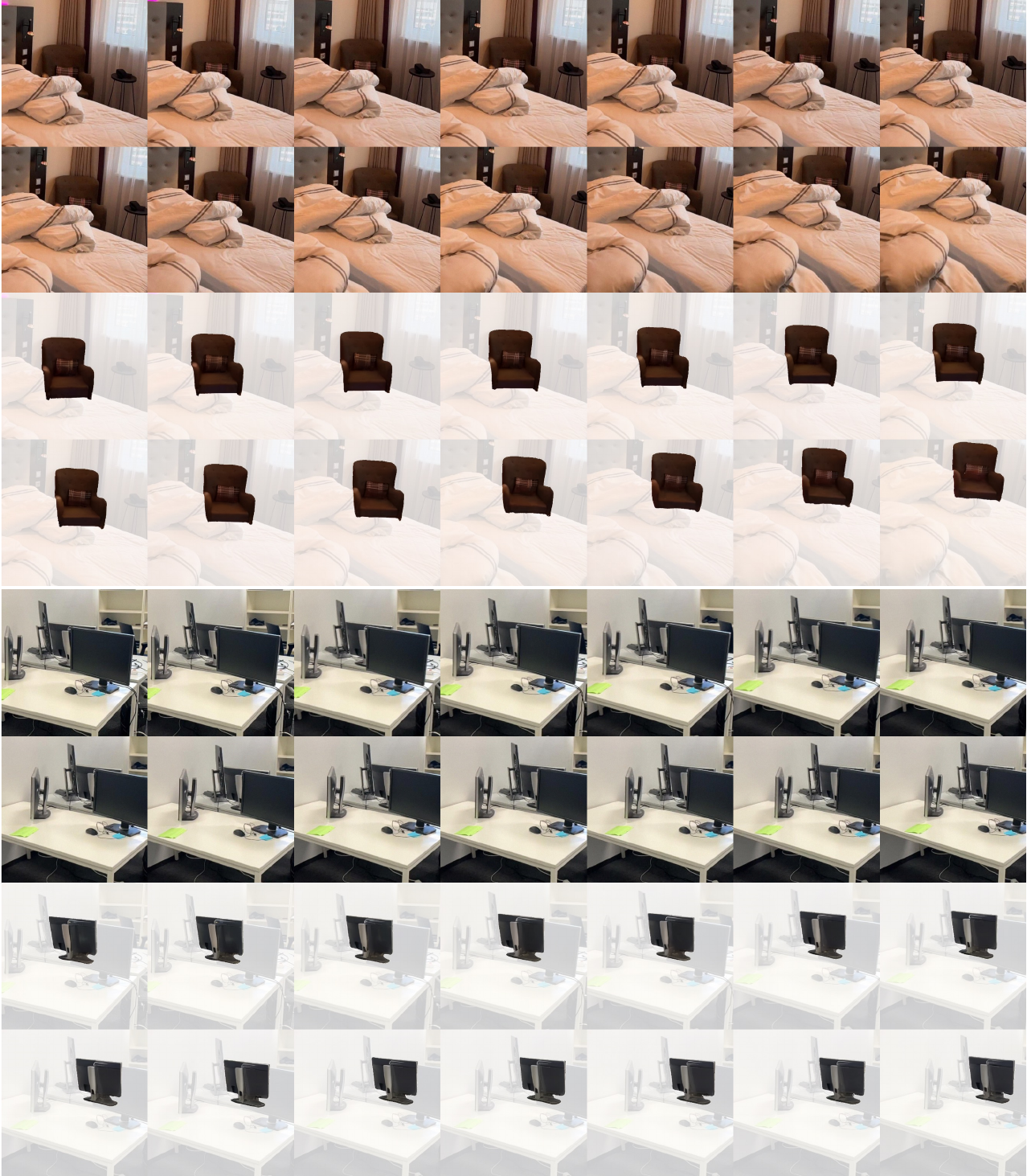}
    \caption{\textbf{Qualitative results on ScanNet~\cite{dai2017scannet, yeshwanth2023scannet++}.}}
    \label{fig:quali_supp_0}
\end{figure*}

\begin{figure*}[t]
    \centering
    \includegraphics[width=\linewidth]{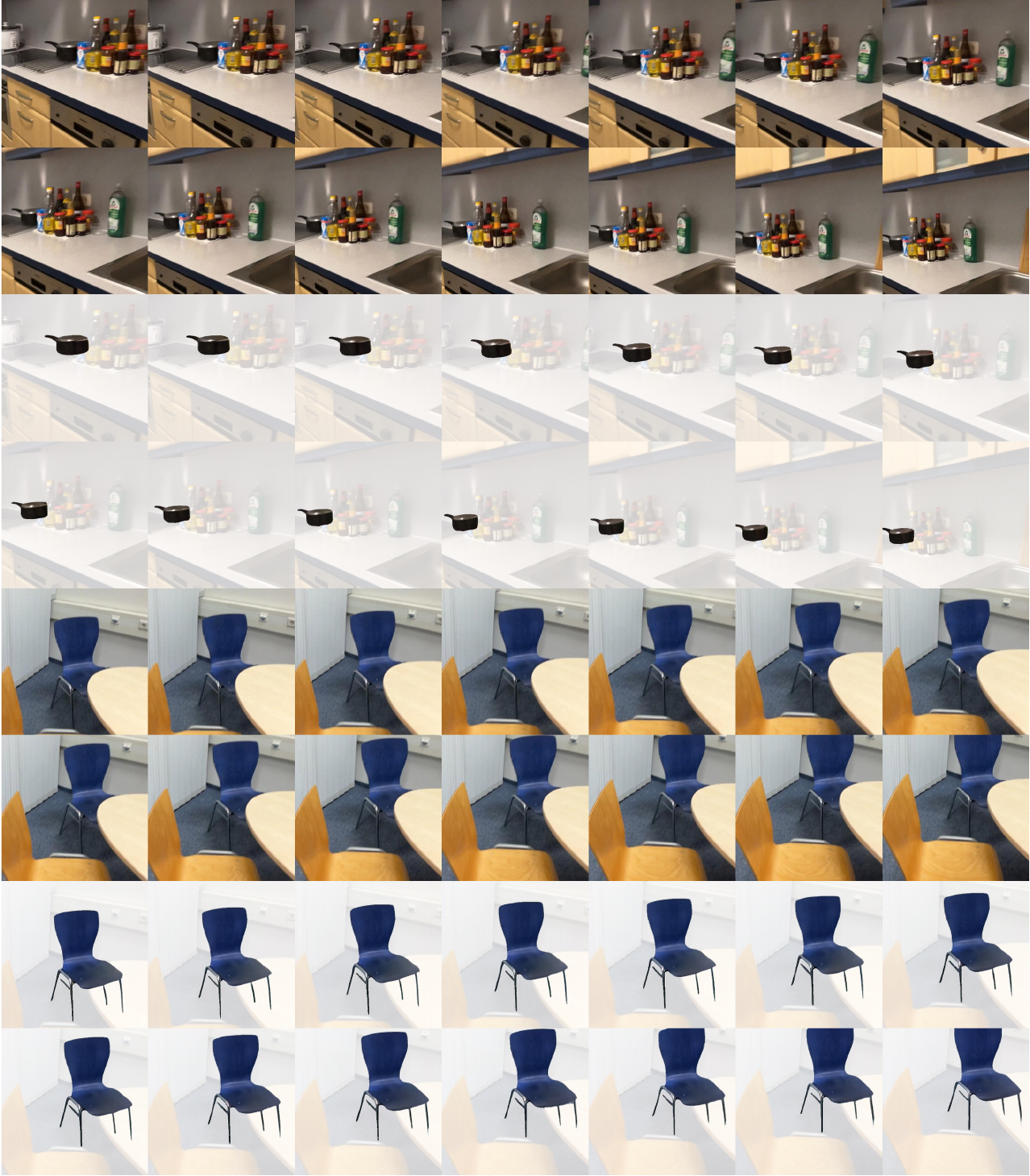}
    \caption{\textbf{Qualitative results on ScanNet~\cite{dai2017scannet, yeshwanth2023scannet++}.}}
    \label{fig:quali_supp_1}
\end{figure*}

\begin{figure*}[t]
    \centering
    \includegraphics[width=\linewidth]{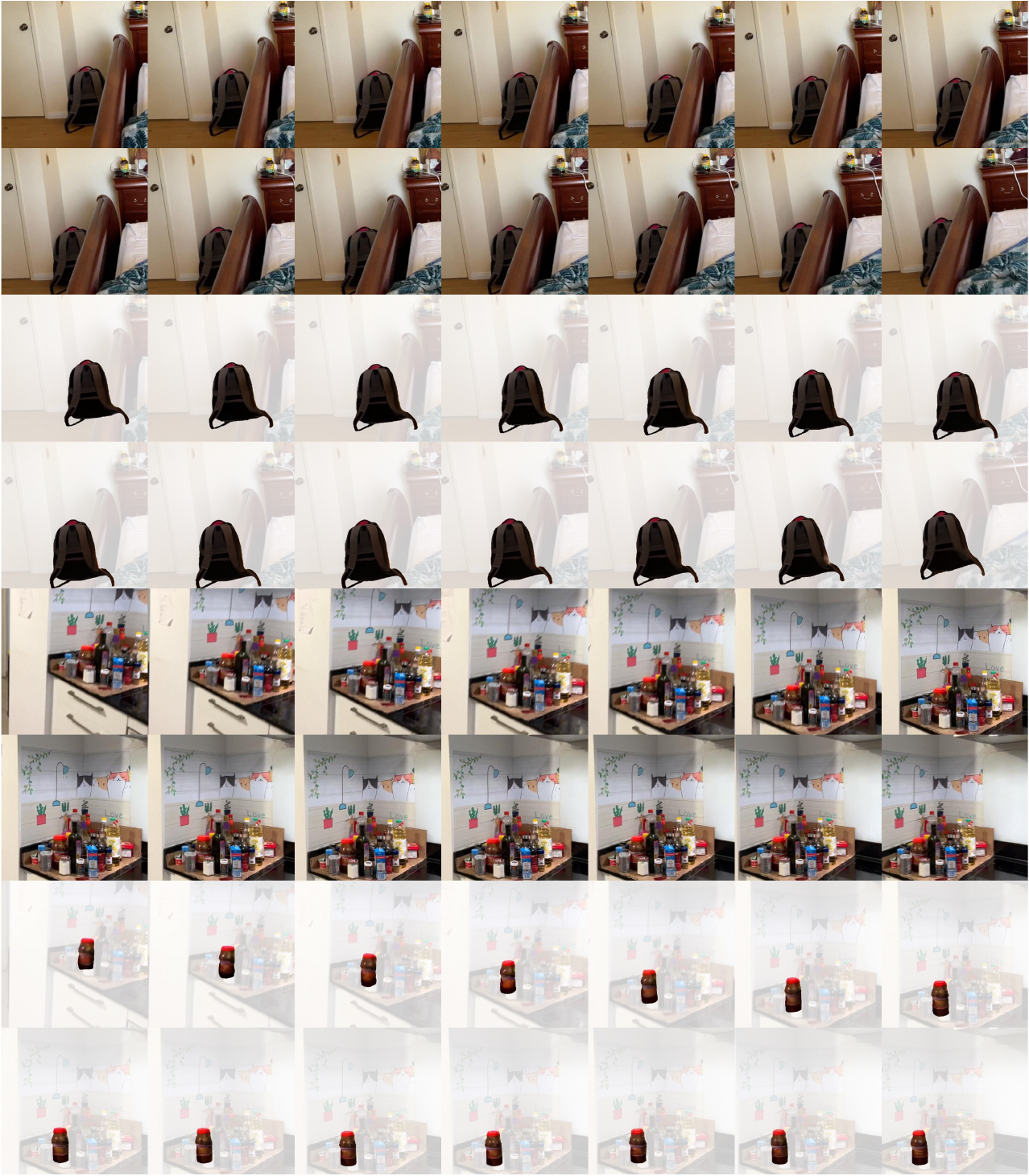}
    \caption{\textbf{Qualitative results on ScanNet~\cite{dai2017scannet, yeshwanth2023scannet++}.}}
    \label{fig:quali_supp_2}
\end{figure*}

\begin{figure*}[t]
    \centering
    \includegraphics[width=\linewidth]{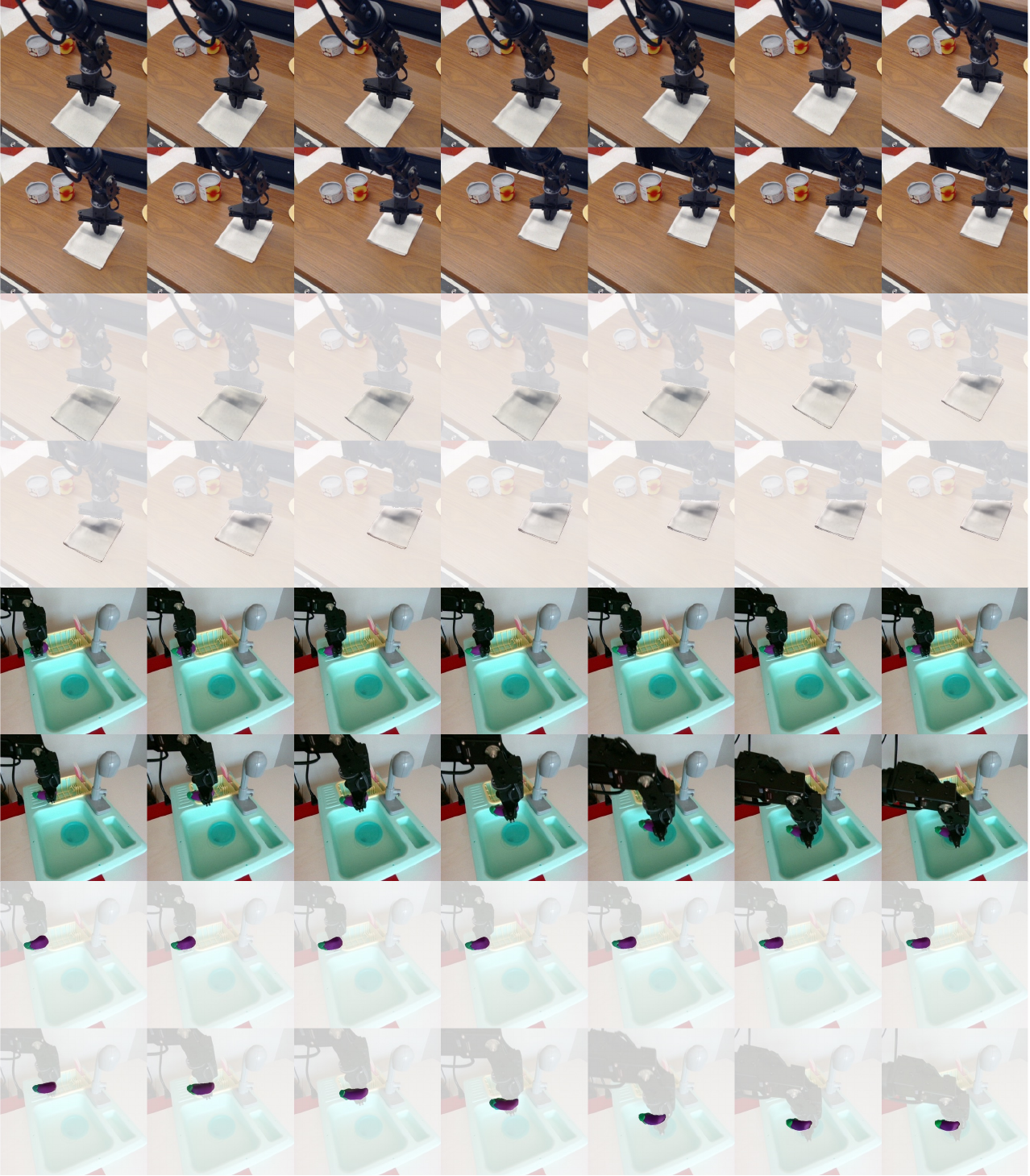}
    \caption{\textbf{Qualitative results on BridgeData~\cite{ebert2021bridge, walke2023bridgedata}.}}
    \label{fig:quali_supp_3}
\end{figure*}

\begin{figure*}[t]
    \centering
    \includegraphics[width=\linewidth]{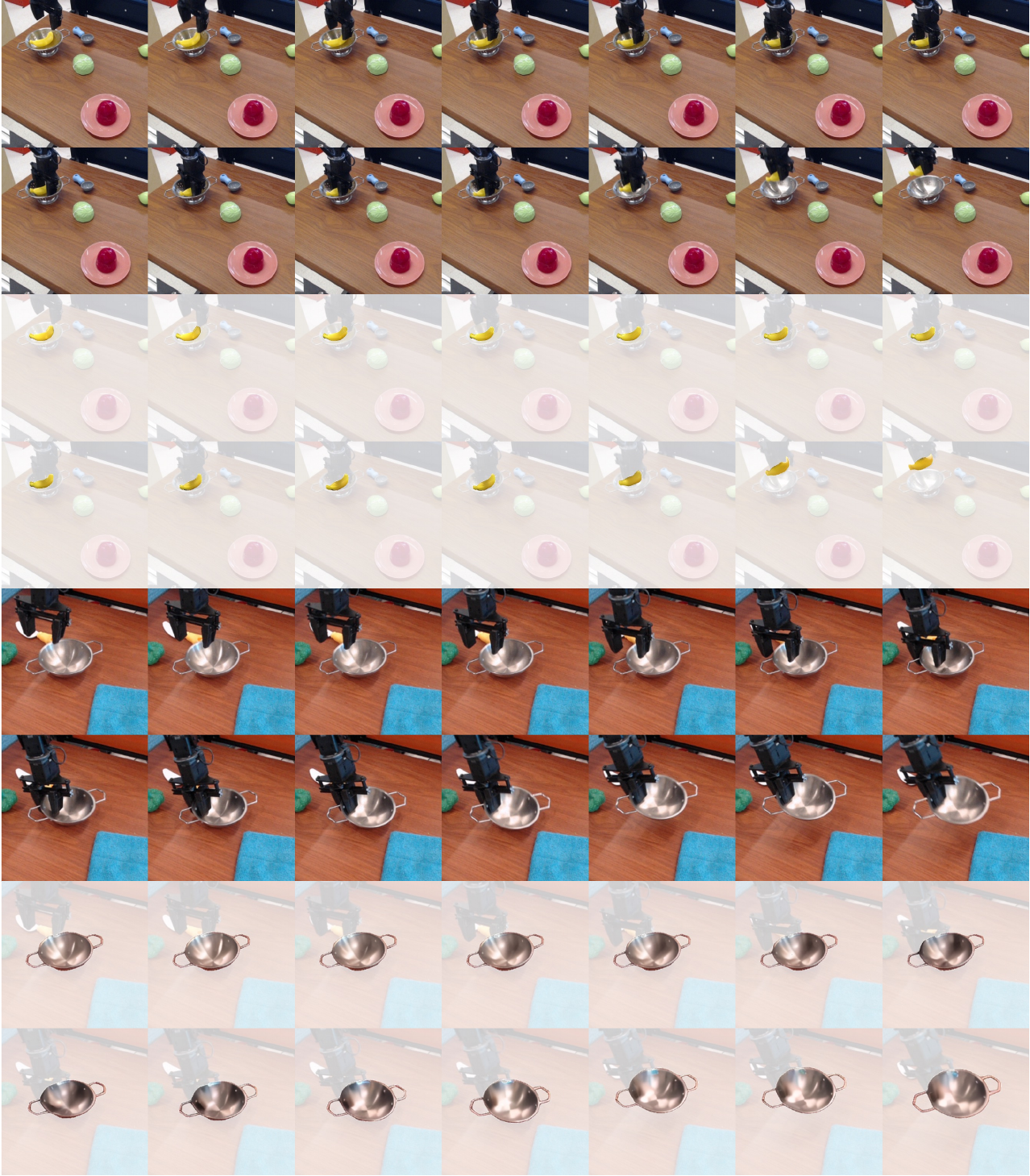}
    \caption{\textbf{Qualitative results on BridgeData~\cite{ebert2021bridge, walke2023bridgedata}.}}
    \label{fig:quali_supp_4}
\end{figure*}

\begin{figure*}[t]
    \centering
    \includegraphics[width=\linewidth]{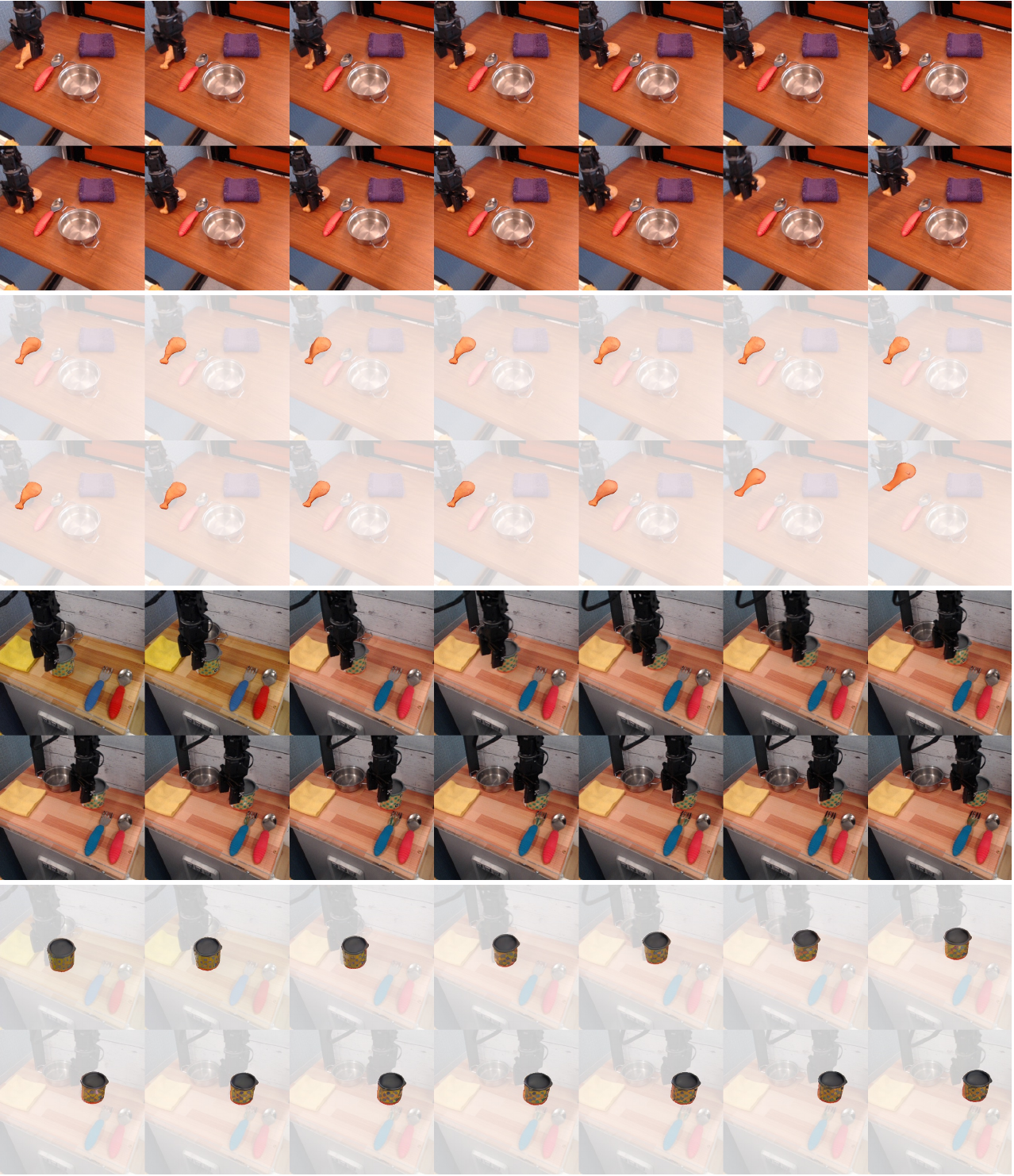}
    \caption{\textbf{Qualitative results on BridgeData~\cite{ebert2021bridge, walke2023bridgedata}.}}
    \label{fig:quali_supp_5}
\end{figure*}

\begin{figure*}[t]
    \centering
    \includegraphics[width=\linewidth]{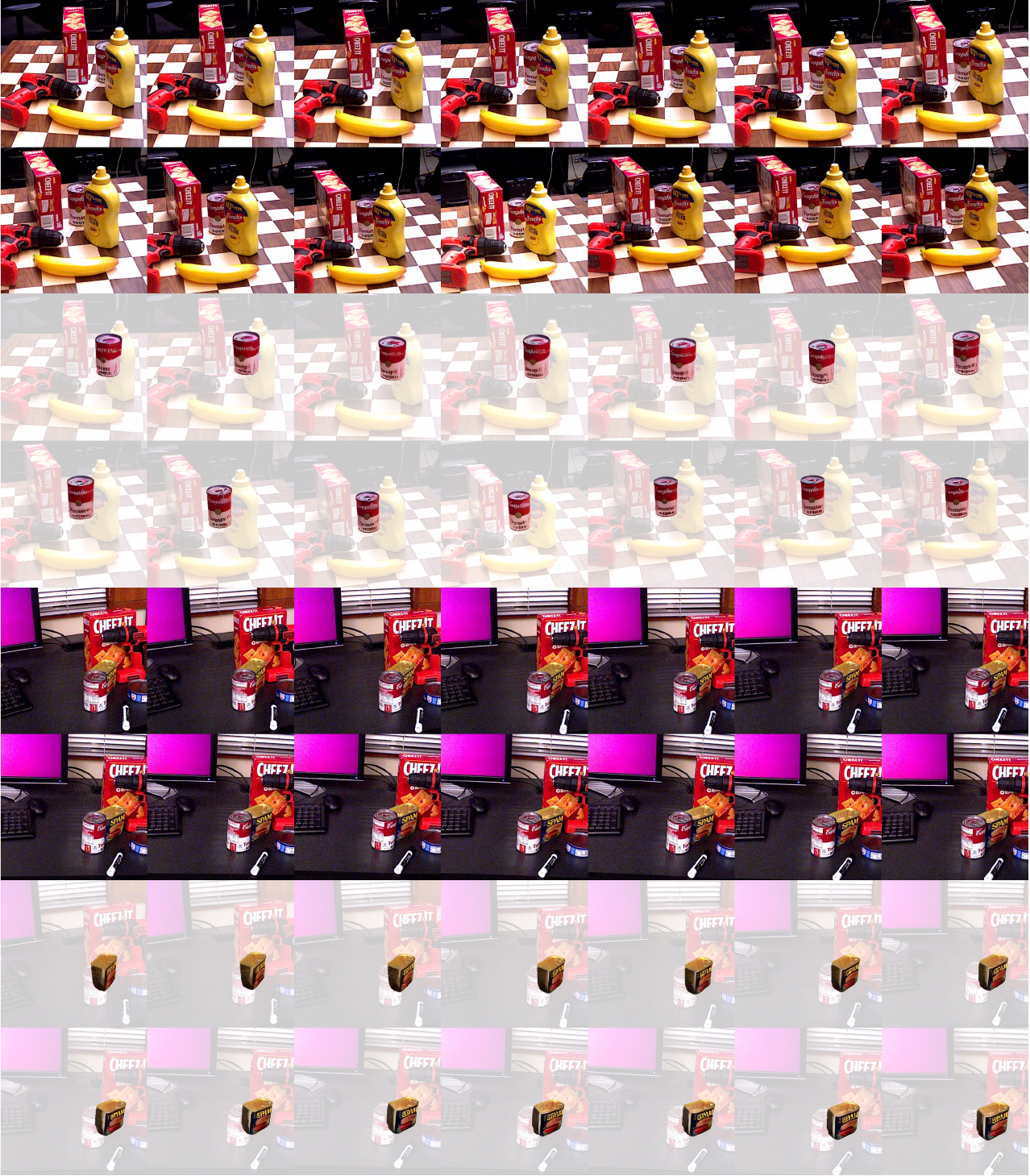}
    \caption{\textbf{Qualitative results on YCB-Video~\cite{xiang2017posecnn}.}}
    \label{fig:quali_supp_6}
\end{figure*}

\begin{figure*}[t]
    \centering
    \includegraphics[width=\linewidth]{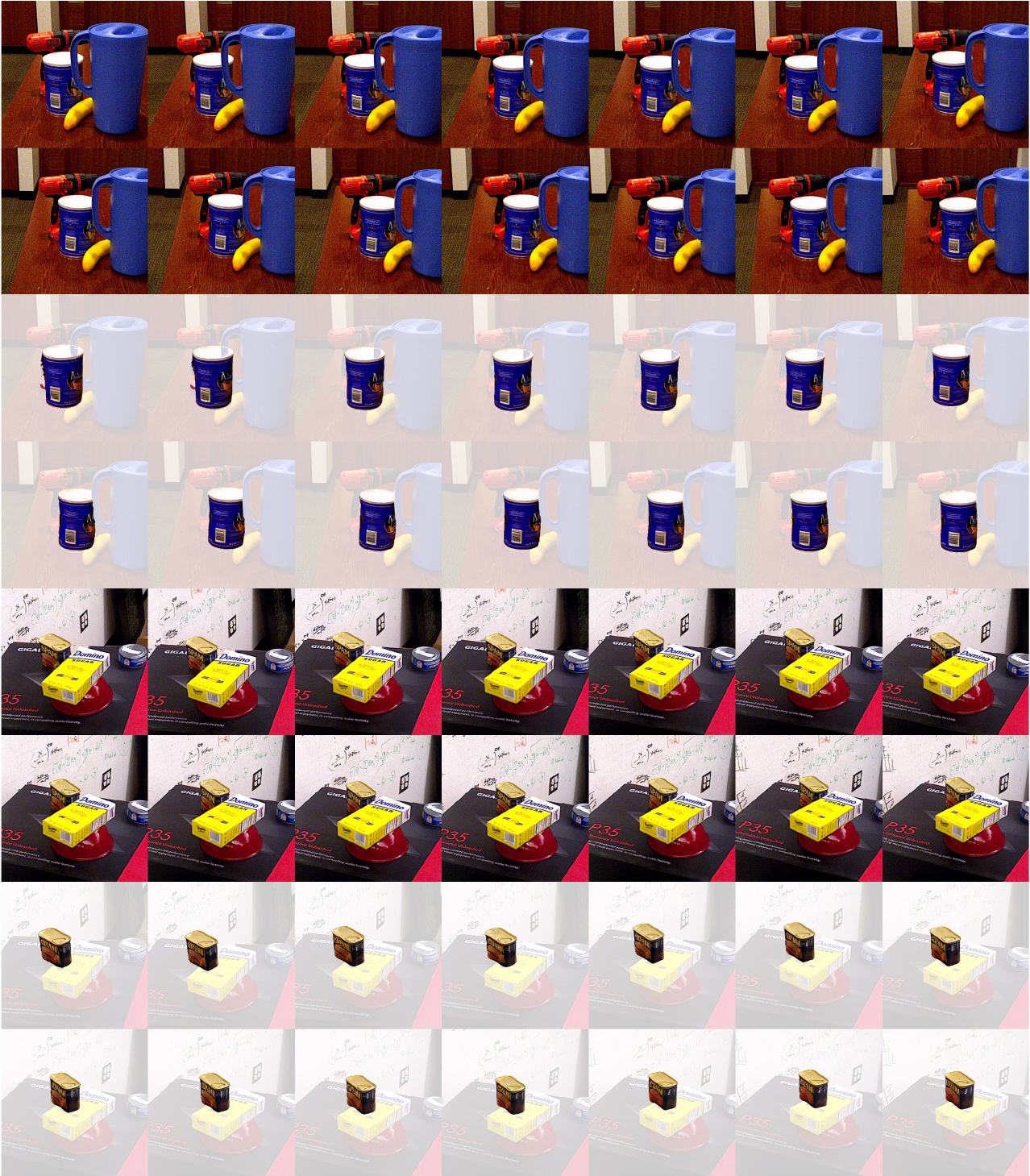}
    \caption{\textbf{Qualitative results on YCB-Video~\cite{xiang2017posecnn}.}}
    \label{fig:quali_supp_7}
\end{figure*}

\begin{figure*}[t]
    \centering
    \includegraphics[width=\linewidth]{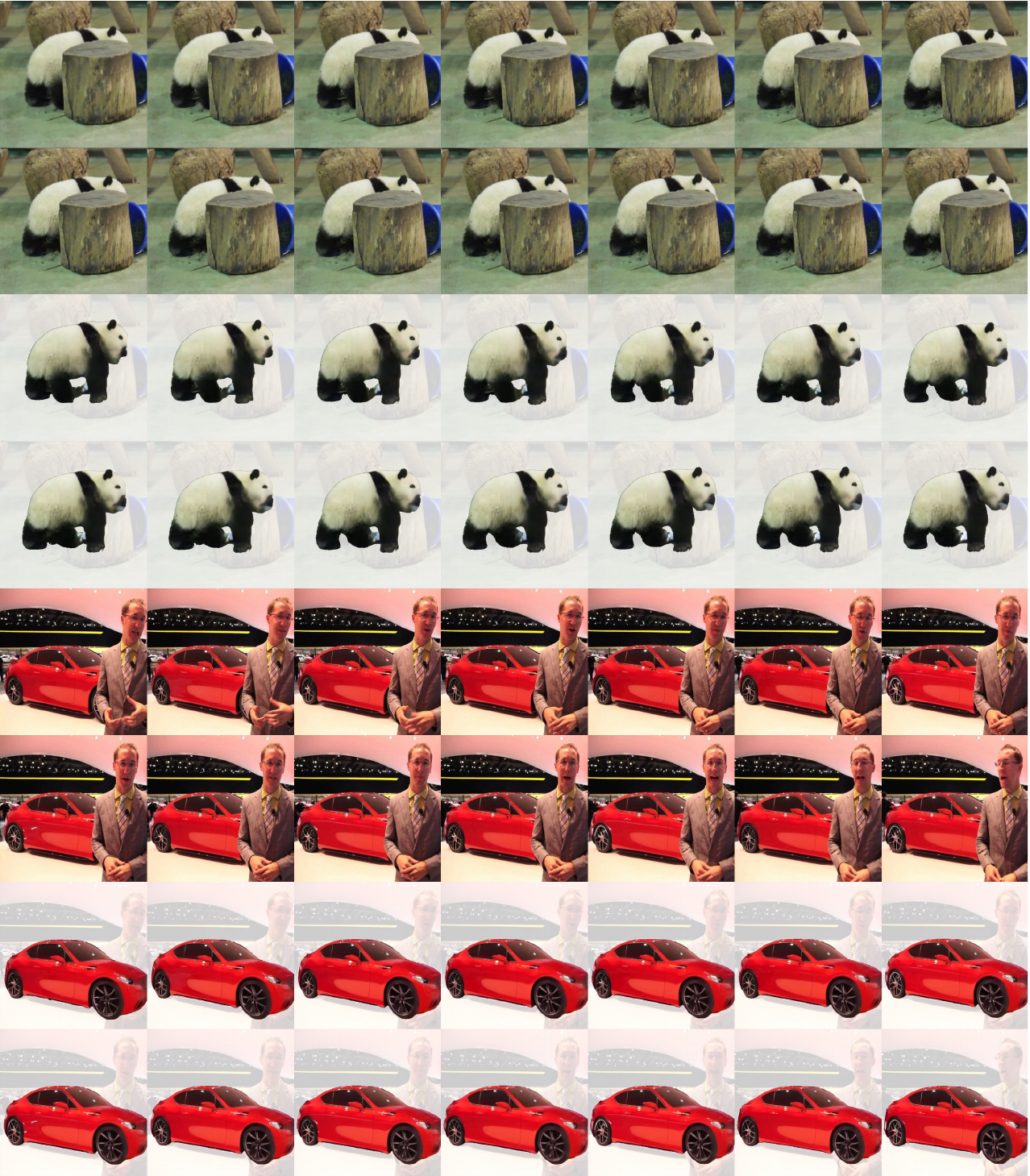}
    \caption{\textbf{Qualitative results on YouTube-VOS~\cite{xu2018youtube}.}}
    \label{fig:quali_supp_8}
\end{figure*}

\begin{figure*}[t]
    \centering
    \includegraphics[width=\linewidth]{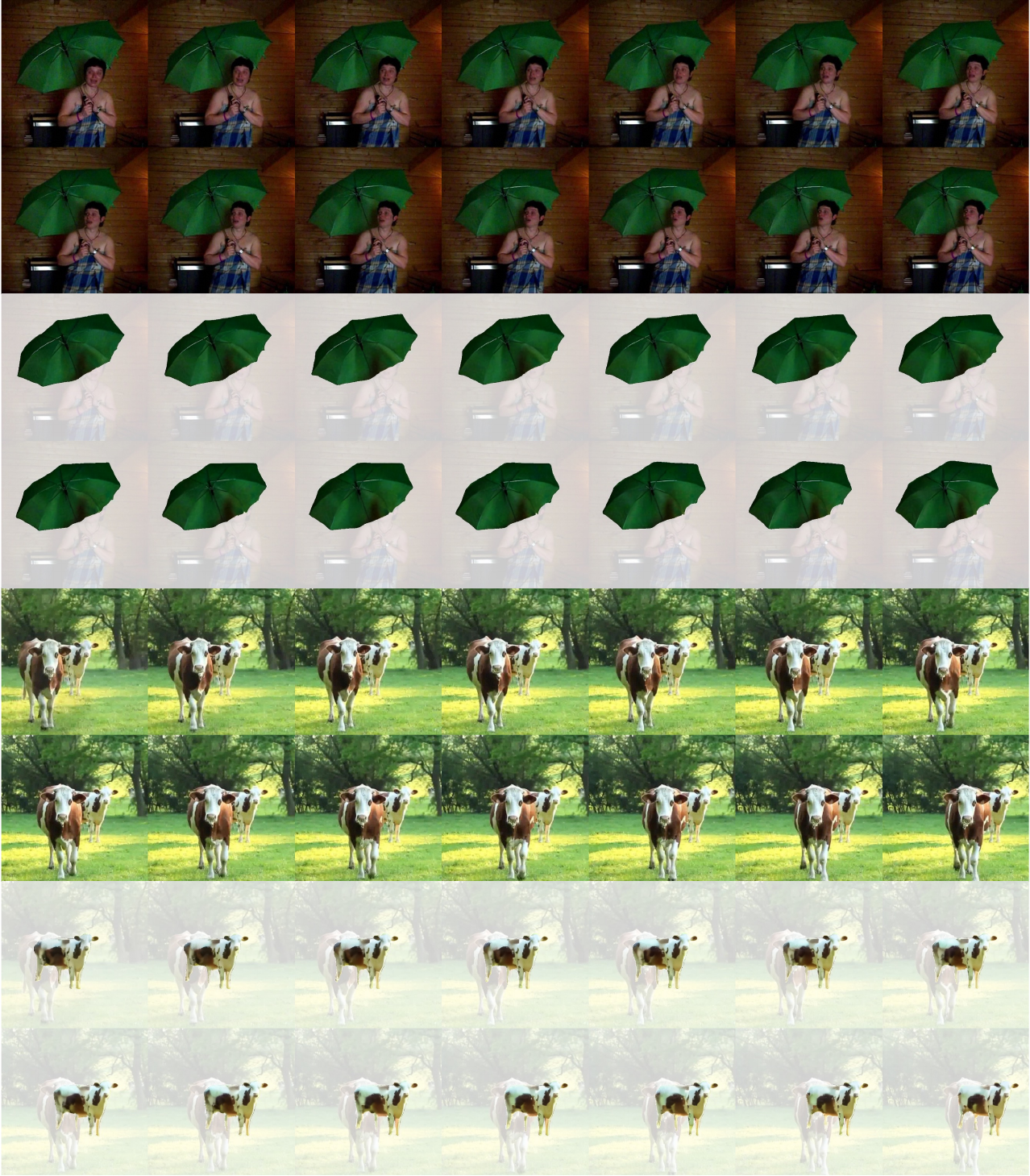}
    \caption{\textbf{Qualitative results on YouTube-VOS~\cite{xu2018youtube}.}}
    \label{fig:quali_supp_9}
\end{figure*}

We provide more qualitative results of our method on on ScanNet~\cite{dai2017scannet, yeshwanth2023scannet++} in \cref{fig:quali_supp_0}, \cref{fig:quali_supp_1}, and \cref{fig:quali_supp_2}. More qualitative results on  BridgeData~\cite{ebert2021bridge, walke2023bridgedata} are provided in \cref{fig:quali_supp_3}, \cref{fig:quali_supp_4}, and \cref{fig:quali_supp_5}. More qualitative results on YouTube-VOS~\cite{xu2018youtube} are provided in \cref{fig:quali_supp_6} and \cref{fig:quali_supp_7}. More qualitative results on YCB-Video~\cite{xiang2017posecnn} are provided in \cref{fig:quali_supp_8} and \cref{fig:quali_supp_9}. More intuitive visualized results on various datasets and in-the-wild videos are provided in the local website attached in \supp.
\subsection{Revealing Missing Objects}
An object may be completely invisible (occluded by other objects) in a video clip. We observe that our method is able to reveal completely missing objects by aggregating information from neighboring frames in certain cases. For example, as shown in \cref{fig:missing}, the black car is completely occluded by the white car in front of it in the middle frames, and our method is able to hallucinate the position, shape, and appearance of the black car. 
\subsection{Tackling Long Videos}
To extend beyond 14 frames, we introduce a sliding window mechanism for progressively generating subsequent frames, as illustrated in \cref{fig:long}. In the first run, the initial 14 frames are selected, and the amodal object is synthesized. The synthesized object is then blended back into the video, as shown in the second row. For the second run, subsequent frames are concatenated with the previously blended frames as the input video. The visible masks of the overlap frames are acquired by thresholding the synthesized object. This sliding window approach allows our method to effectively generate videos exceeding 14 frames. Two additional qualitative examples are available on the local website.
\subsection{Diversity in Sampling}
Since amodal completion possesses inherent ambiguity, we can synthesize multiple reasonable results, and a qualitative example indicating the diversity is shown in \cref{fig:diversity}.
\subsection{Failure Cases}
\label{sec:fail}
We illustrate several failure cases in \cref{fig:fail}. In the first example, nearly the entire legs and arms of the human are occluded, representing a severely occluded scenario. Under such conditions, our method may produce low-quality outputs, such as blurry hands and occasionally missing legs. In the second example, the chair is heavily occluded by the white table. While it can be inferred that the chair has thin legs, our method may struggle with accurately reconstructing thin structures in certain frames. In the third example, the camera undergoes drastic movements, and the brown chair becomes heavily occluded from specific viewpoints. Our method struggles to recover accurate and consistent results when the object is almost missing.
\section{Limitations and Negative Impacts}
\paragraph{Limitations} We have discussed about several failure cases in \cref{sec:fail}. Additionally, our method is sensitive to resolution variations due to the constraints of the SVD architecture. While we can extend beyond 14 frames, generating extremely long videos remains a challenge. We hope that advancements in more powerful video diffusion models will help address these issues in the future. Moreover, as our method incorporates data from only a few datasets, as mentioned in \cref{sec:dataset}, its performance may degrade when generalizing to vastly different domains, such as human interactions with other objects, as shown in the example in \cref{fig:fail}. Incorporating more diverse training data and curating realistic occlusion scenarios could help mitigate this issue.
\paragraph{Negative Impacts} The use of diffusion models to generate content raises significant ethical concerns, including potential privacy violations and the risk of generating biased content. These models can be misused to spread misinformation or serve deceptive purposes, eroding trust and causing societal harm. Additionally, they may produce misleading or false outputs, causing potential challenges in fields where accuracy and reliability are crucial.

\clearpage



\end{document}